\documentclass[runningheads]{llncs}

\usepackage{eccv}

\usepackage{eccvabbrv}

\usepackage{graphicx}
\usepackage{booktabs}
\usepackage{multirow}
\usepackage{makecell}
\usepackage[table]{xcolor}
\usepackage{tabularx}

\usepackage[accsupp]{axessibility}  % Improves PDF readability for those with disabilities.

\usepackage{hyperref}

\usepackage{orcidlink}

\definecolor{avgpurple}{HTML}{E6E6FF}
\definecolor{avgblue}{HTML}{E7EFFA}
\definecolor{avggray}{HTML}{F2F2F2}

\newcommand{\tableref}[1]{%
  \ifstrequal{#1}{tab:logits_comp}{Table \hyperref[tab:logits_comp]{6}}%
  {\ifstrequal{#1}{tab:setcomp}{Table \hyperref[tab:setcomp]{7}}%
  {\tablename~\ref{#1}}}%
}

\begin{document}

\title{Towards Purified Multi-Label Test-Time Adaptation of 
Vision-Language Models} 
% \title{Boosting Multi-Label Test-Time Adaptation of Vision-Language Models via Purification} 

\titlerunning{Towards Purified Multi-Label Test-Time Adaptation}


% \author{ Yiwen Liang\inst{1}\orcidlink{0009-0004-8936-4632} \and Hui Chen\inst{1}\thanks{Corresponding author.} \orcidlink{0000-0003-4180-5801}\and Yizhe Xiong\inst{1}\orcidlink{0009-0001-5233-9466} \and Mengyao Lyu\inst{1}\orcidlink{0000-0002-5404-4127} \and Yuhan Cao\inst{3}\orcidlink{0009-0008-7569-519X} \and Zijia Lin\inst{1}\orcidlink{0000-0002-1390-7424} \and Shuaicheng Niu\inst{2}\orcidlink{0000-0001-8212-1831} \and Sicheng Zhao\inst{1}\orcidlink{0000-0001-5843-6411} \and Jungong Han\inst{1}\orcidlink{0000-0003-4361-956X} \and Guiguang Ding\inst{1}\textsuperscript{$\star$}\orcidlink{0000-0003-0137-9975}  }
\author{ Yiwen Liang\inst{1} \and Hui Chen\inst{1}\thanks{Corresponding author.} \and Yizhe Xiong\inst{1} \and Mengyao Lyu\inst{2} \and Yuhan Cao\inst{3} \and Zijia Lin\inst{1} \and Shuaicheng Niu\inst{2} \and Sicheng Zhao\inst{1} \and Jungong Han\inst{1} \and Guiguang Ding\inst{1}\textsuperscript{$\star$} }

% \author{First Author\inst{1}\orcidlink{0000-1111-2222-3333} \and Second Author\inst{2,3}\orcidlink{1111-2222-3333-4444} \and Third Author\inst{3}\orcidlink{2222--3333-4444-5555}}

\authorrunning{Liang et al.}

\institute{Tsinghua University, Beijing, China \and Nanyang Technological University, Singapore \and University of Washington, Seattle, WA, USA\\ 
% \email{ \{evenliang789,jichenhui2012,schzhao,jungonghan77\}@gmail.com, xiongyizhe2001@163.com, mengyao.lyu@outlook.com, linzijia07@tsinghua.org.cn, shuaicheng.niu@ntu.edu.sg, dinggg@tsinghua.edu.cn } 
\email{ \{evenliang789,jichenhui2012\}@gmail.com, dinggg@tsinghua.edu.cn } 
}
% Springer Heidelberg, Tiergartenstr.~17, 69121 Heidelberg, Germany

\maketitle

\begin{abstract}
Test-time adaptation (TTA) has been widely explored in single-label recognition, effectively mitigating distribution shifts, especially when combined with vision-language models. 
However, real-world images often contain multiple objects, while the more practical multi-label test-time adaptation (MLTTA) has received little attention so far.
% Conventional TTA methods exhibit inherent limitations in multi-label scenarios, as they rely on global image-level representations that blend information from multiple co-occurring objects, leading to inter-class feature entanglement that impairs class-specific discriminability and adaptation effectiveness.
Recent cache-based TTA methods have shown promising efficiency and effectiveness, yet directly extending them to multi-label scenarios suffers from a one-to-many mapping problem: a shared global representation entangling co-occurring objects is stored as class-wise cache prototypes, inducing dominant-label bias and compromised cache calibration.
% However, directly applying them to multi-label scenarios suffers from a one-to-many mapping problem: a shared global image-level representation that entangles co-occurring objects within an image is stored as class-wise cache prototypes, leading to dominant-label bias and compromised cache calibration.
% that blends information from multiple co-occurring objects is stored as class-wise cache prototypes, leading to dominant-label bias, weakened inter-class discriminability, and compromised cache calibration.
% , inducing dominant-label bias and compromising cache calibration.
% a shared global image-level representation may simultaneously align with multiple objects, inducing dominant-label bias and compromising cache calibration.
% 此外，没能很好应用 exploit the information Fine-grained region evidance 
% In addition, they fail to effectively leverage fine-grained region evidence, since both region selection and utilization remain non-trivial.
While introducing region-level cues helps isolate class-specific evidence, such regional evidence can also be unreliable under distribution shifts, making its identification and utilization non-trivial.
% A natural remedy is to introduce region-level cues to better isolate class-specific evidence. However, region evidence under distribution shifts can also be unreliable, and its selection and utilization remain non-trivial.
To address these issues, we introduce \textbf{PuRF}, a novel \textbf{PuR}i\textbf{F}ication-driven cache-based method for multi-label test-time adaptation of vision-language models.
% RePT (TTA with \textbf{Pur}i\textbf{f}ication, a novel cache-based method for multi-label test-time adaptation of vision-language models via region-aware purification.
% PuTTA introduces a dedicated decoupling and region-based caching mechanism, which performs informative region selection and effective feature integration to enhance class-wise discriminability and leverage comprehensive region information.
% Specifically, Purf performs reliable region selection and integration to exploit comprehensive region cues and enhance fine-grained alignment, and further constructs a discriminative region-based cache to improve adaptation.
Specifically, PuRF first performs region purification to identify reliable regions, providing comprehensive regional cues for multi-label recognition and enabling fine-grained alignment.
Based on these purified regions, PuRF conducts cache purification to enhance cache representation and adaptability, where episodic purification builds a discriminative region-based cache, and temporal refreshing further promotes long-term cache adaptability.
% further builds a discriminative region-based cache for cache purification.
% Furthermore, to mitigate the rapid saturation caused by multiple entries per multi-label image, we design an adaptive refreshing mechanism that promotes the temporal adaptability of cache.
% Furthermore, to mitigate early-stage bias and rapid cache saturation caused by multiple entries in multi-label images, we introduce temporal purification to promote long-term adaptability of cache.
% In addition, to prevent cache saturation caused by multiple cached entries per multi-label image, we design an adaptive cache refreshing mechanism to promote the temporal adaptability of cache.
% Experiments on multi-label benchmarks demonstrate that PuRF outperforms state-of-the-art methods by 3.37\% mAP on average with superior effectiveness and better generalization. 
Experiments demonstrate that PuRF consistently outperforms state-of-the-art methods, achieving a notable 4.05\% mAP improvement on ViT-B/32 across five datasets.
% The code is available at: \url{https://github.com/Evelyn1ywliang/PuRF-MLTTA}.
% Codes will be available publicly.
\end{abstract}

%%%%%%%%%%%%%%%%%%%%%%%%%%%%%%%%%%%%%%%%%%%%%%%%%%%%
\section{Introduction}
\label{sec:intro}
% 这一段已经写了vlm在single, multi-label下面的能力, 然后接下来又写了test-time下面遇到的问题.
Vision-Language Models (VLMs)~\cite{clip,align} have exhibited remarkable zero-shot capability, successfully tackling a wide range of computer vision tasks~\cite{lee2025ratta,raclip_retrieval,retrievetta2,stata}.
% including classification~\cite{HSPNet,darprompt}, retrieval~\cite{lee2025ratta,raclip_retrieval,retrievetta2}, and segmentation~\cite{rpn_seg,consolidator}. 
However, these models often fail to generalize well when facing testing distributions that deviate from those seen during pre-training, known as distribution shift~\cite{cdisvit,learnfrom,pyra}. Such shifts commonly occur in real scenarios, such as images captured by different environments~\cite{environ,domainNet} or sensing conditions~\cite{imgC,objectNet,NAVIA}.

\begin{figure}[!t]
% \centering
%     % \vspace{-0.3cm}
% \centerline{\includegraphics[width=7.3cm]{figure/pseudo-fig1.png}} 
%     \vspace{-0.3cm}
% \caption{Systematic comparison of three TTA paradigms. Prompt-based methods suffer from inefficiency caused by episodic resets, whereas cache-based approaches remain noisy and biased due to accumulated global feature noise and dominance bias. In contrast, our ReD-MLTTA achieves efficient and accurate multi-label adaptation via region-disentangled and reliable caching.}
%     \label{fig: 1}
%     \vspace{-0.5cm}
    \centering
    % ---------- 第一行子图 ----------
    % \begin{subfigure}{0.95\columnwidth}
    %     \hspace*{-0.58cm} % 负值左移
    %     \centering
    %     \includegraphics[width=\linewidth]{figure/fig1_existing_1.pdf}
    %     \vspace{-2pt} %
    %     \caption{Existing cache-based methods}
    %     \label{fig:cache_based}
    % \end{subfigure}
    
    % % \vspace{5pt} % 两张图之间的垂直间距
    
    % % ---------- 第二行子图 ----------
    % \begin{subfigure}{0.99\columnwidth}
    %     \centering
    %     \includegraphics[width=\linewidth]{figure/fig1_ours_1.pdf}
    %     \caption{Region-disengangled method for multi-label TTA (Ours)}
    %     \vspace{-2pt} %
    %     \label{fig:ours_method}
    % \end{subfigure}

    \begin{minipage}[b]{1.0\linewidth}
		\centering
        \vspace{-0.08cm}
		\centerline{\includegraphics[width=11.5cm]{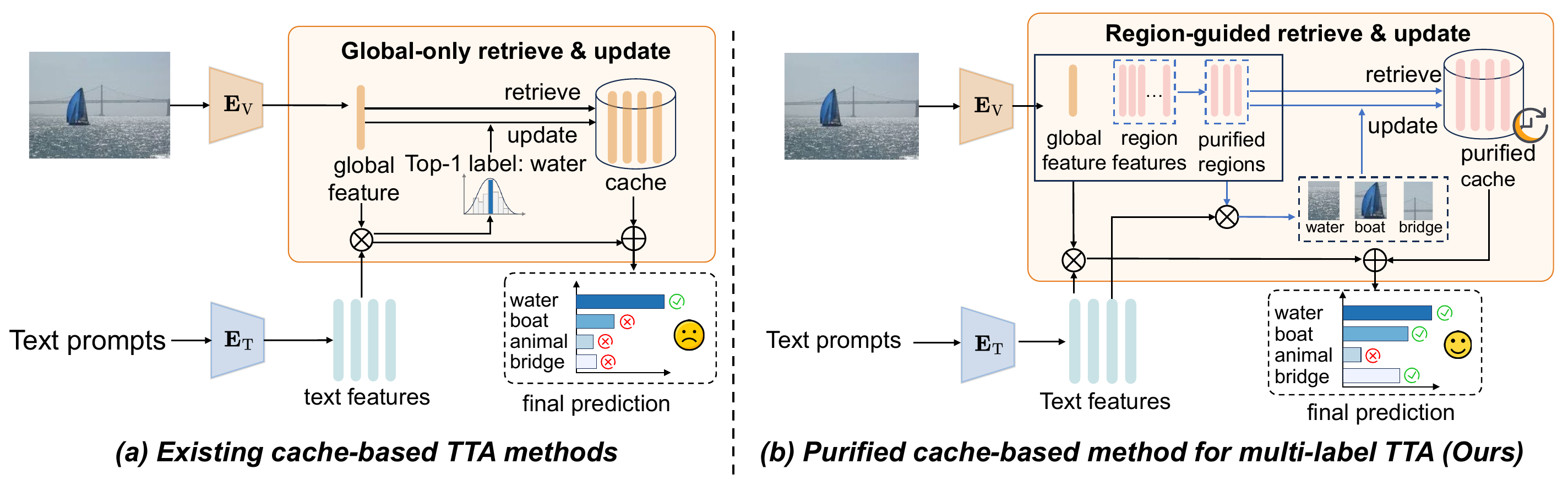}}
		 \vspace{-0.23cm}
	\end{minipage}

    % ---------- 总标题 ----------
    \caption{
    Comparison between existing cache-based methods and ours.
    % Paradigm comparison.
    (a) Existing cache-based TTA methods~\cite{tda,dpe,reta} suffer from one-to-many representation coupling in multi-label TTA, where shared global features are dominated by salient classes (e.g., ``water''), misleading class-wise cache calibration.
    % 会导致被 主导类别误导 ("boat") 而 cache 的xxx 失效. 
    %feature entanglement and suboptimal alignment in multi-label TTA, as global features cannot serve as reliable class-wise prototypes for caching.
    (b) PuRF performs region purification and cache purification, enabling fine-grained and accurate multi-label adaptation.
    % constructs a refined region-based cache and achieves fine-grained recognition.
    }
    \vspace{-0.54cm}
    \label{fig:1}
\end{figure}

% p2: TTA, but most focus on single label
Test-time adaptation (TTA) has emerged as a practical solution for handling distribution shifts by adapting models at inference using only unlabeled test data. 
Recent studies primarily leverage learnable prompts~\cite{tpt,difftpt,scap} for dynamic alignment and cache mechanisms~\cite{tda,cosmic,multicache, reta} for historical knowledge retrieval to enhance the adaptation capability of VLMs.
However, most existing methods are developed for \textit{single-label recognition}, whereas real-world images often contain multiple co-occurring objects, rendering \textit{multi-label test-time adaptation} substantially more challenging and complex than single-label settings.
% Under such a multi-label scenario, these approaches~\cite{swapprompt,promptalign} typically rely on entropy minimization, which tends to produce spiky prediction distributions that overemphasize a single dominant label while suppressing others~\cite{memo,eata}, making the \textit{multi-label test-time adaptation} considerably more complex than in single-label setting.

% p3: ML-TTA作为第一个解决这个方案的东西， xxxx
As the first work on multi-label TTA, ML-TTA~\cite{mltta} introduces a Bound Entropy Minimization objective with prompt optimization to enhance the confidence of top-$k$ labels while mitigating over-suppression from naive entropy minimization.
% 但是还存在了什么问题: episodic prompt resetting, inefficient. 
% However, its adaptation process involves frequent text corpus retrieval, episodic prompt resetting, and back-propagation through the entire encoder, leading to \textit{computational inefficiency}.
However, it still relies on intensive text retrieval and episodic prompt resetting, 
% failing to effectively exploit historical information to assist prediction and adaptation, leading to suboptimal performance.
failing to exploit historical information and resulting in suboptimal performance.
In contrast, cache-based methods~\cite{dpe,reta,multicache,boostadapter} have demonstrated strong performance in single-label TTA by accumulating and reusing historical information for prediction refinement via retrieval-based aggregation, but remain largely unexplored in multi-label settings, leaving their potential underexploited.
% maintaining representative class-specific prototypes for prediction refinement via retrieval-based aggregation, 
% 没有体现一张图片里面的 one vs. all的情况, 以及没有体现用全局特征也用来检索的问题
% If directly applied to multi-label images will raise a critical question: they typically store one spatially pooled global representation as class-wise prototypes, which leads to feature entanglement due to the mixture of co-occurring object features in shared embeddings.
%one-vs-many dilemma
When directly applied to multi-label images, these methods encounter a \textit{one-vs-many dilemma}~\cite{OFOL1,CSRA}, where a single spatially pooled global feature represents all co-occurring objects and simultaneously acts as a shared prototype for cache retrieval and storage.
%dominated by salient objects
Such shared representations cannot effectively capture label-specific features~\cite{OFOL1} and tend to be dominated by salient objects~\cite{DCLIP, Dual-View, MuP-VSS}, 
leading to biased predictions and weakened label-specific discriminability.
%cache calibration unreliable
Consequently, the cache fails to perform reliable class-specific calibration, thereby hampering adaptation performance, as illustrated in Fig.~\ref{fig:1}\hyperref[fig:1]{(a)}. 
%region-level methods exist
Although prior works~\cite{lesa, TRM-ML, CSRA} incorporate region-level cues to improve class-specific representation in multi-label recognition, 
obtaining reliable region evidence and leveraging it in fully unsupervised test-time settings remains challenging, as naive region proposals are often noisy and include irrelevant context~\cite{lsnet,RAM,Query-Based-Knowledge}.

% \textcolor{red}{leading to \textit{feature entanglement} in shared representations.
% Feature entanglement~\cite{entangle1,entangle2} refers to the phenomenon where features of co-occurring objects overlap and interfere with each other in the representation space. 
% The mixed image-level embeddings fail to represent multiple objects and exhibit weakened class-specific discriminability, thus hampering adaptation performance, as shown in Fig.~\ref{fig:1}\hyperref[fig:1]{(a)}. 
% }
% While cache-based methods alleviate these limitations by efficiently storing and retrieving features without costly retraining, they remain largely unexplored in multi-label settings and usually store global representations as class-wise prototypes, which can cause \textit{feature entanglement} due to the mixture of co-occurring object features in shared representations.

% The resulting collapsed 

% with weakened class-specific discriminability, thus hampering adaptation performance, as shown in Fig.~\ref{fig:1}\hyperref[fig:1]{(a)}. 
% blurring class boundaries and leading to inaccurate recognition (Fig.~\ref{fig:1}\hyperref[fig:1]{(a)}).}

% p4: 我们的方法
After revealing the above limitations,
we propose \textbf{PuRF}, a \textbf{PuR}i\textbf{F}ication-driven cache-based method for multi-label Test-Time Adaptation, which builds upon the cache mechanism for its effectiveness with historical information while extending this paradigm to more practical multi-label scenario.
Unlike ML-TTA~\cite{mltta} that tailors optimization objectives for multi-label settings, 
our key idea is to exploit purified regional evidence to enable finer-grained multi-label recognition and construct a more discriminative cache to better utilize historical information.
% through selective region purification, thereby enabling more accurate multi-label recognition and improving the discriminability of class-specific cache calibration.
% \textcolor{red}{our key idea is to better leverage cached historical knowledge to advance fine-grained alignment for accurate recognition and mitigate feature entanglement.}
% enable more fine-grained alignment for accurate multi-label recognition by mitigating feature entanglement.
To this end, PuRF proposes two main components that collectively enhance multi-label adaptation: 
1) Region Purification (RP) via multi-granularity consistency identifies purified regions via relative activation-based selection and derives reliable pseudo supervision via global-local consistency, enabling fine-grained alignment and effective semantic optimization for accurate multi-label test-time recognition;
% refining semantic distributions via residual optimization, achieving more precise recognition by integrating comprehensive semantic information;
% leverages global-local predictive consistency to derive reliable pseudo-labels and refines semantic distributions via residual optimization, achieving more precise recognition by integrating comprehensive semantic information; 
% 1) Multi-granularity Consistency Alignment (MCA) leverages global-local predictive consistency to derive reliable pseudo-labels and refines semantic distributions via residual optimization without full back-propagation, achieving more precise recognition by integrating comprehensive semantic information; 
% 2) Cache Purification (CP) builds class-specific cache prototypes from purified high-confidence regions, forming a cache with more discriminative class representations; 
% 3) Temporal Cache Refreshing (TCR) further enhances cache adaptability under evolving distributions through a time-aware entropy weighting mechanism.
(2) Cache Purification (CP) integrates episodic purification to build class-specific prototypes from purified high-confidence regions for more discriminative representations, and temporal refreshing to improve long-term cache adaptability under evolving distributions via time-aware entropy weighting.
% Overall, ReD provides a robust foundation for reliable and efficient multi-label adaptation, with its effectiveness and superiority consistently demonstrated across five standard multi-label datasets.
% Overall, ReD advances multi-label test-time adaptation by bridging the gap for cache-based methods and extending their applicability from single-label to multi-label recognition, achieving an impressive average gain of xx\% mAP over the strongest state-of-the-art method.}
In Fig.~\ref{fig:1}\hyperref[fig:1]{(b)}, our purified regional evidence and purified cache improve class-specific discriminability for better multi-label recognition.
% Overall, Purf advances multi-label test-time adaptation by 
% extending cache-based paradigm to multi-label settings with two novel strategies, while exhibits strong generalization across diverse backbones and prompt settings with high efficiency, demonstrating its xxx and superiority. 
Overall, PuRF advances multi-label test-time adaptation by extending the cache-based paradigm to multi-label settings with two novel purification strategies. 
Experimental results show that PuRF exhibits strong generalization across diverse backbones and prompt settings while maintaining high efficiency.

% that more effectively leverage historical information, achieving an impressive average gain of 3.37\% mAP over the strongest state-of-the-art method.

% bridging the gap for cache-based methods and 
% 介绍本文提出的方案 (p4)
We summarize our contributions as follows:
\begin{itemize}
% \item We propose ReD, a novel cache-based method for multi-label test-time adaptation that achieves efficient and fine-grained adaptation without requiring full model backpropagation.
\item We propose PuRF, a novel purification-driven cache-based method tailored for multi-label test-time adaptation, enabling fine-grained recognition while enhancing label-specific cache calibration.
% region-disentangled cache-based method tailored for multi-label test-time adaptation, enabling fine-grained recognition while alleviating feature entanglement.
% \item Purf integrates Multi-granularity Consistency Purification (MCP) for adaptive region cue selection and fine-grained global-local alignment,
% Region-aware Cache Purification (RCP) for building class-specific cache prototypes from purified high-confidence regions to enhance class-specific discriminability and reliable calibration;
% and Temporal Cache Refreshing (TCR) for promoting temporal adaptivity into cache updating.
\item PuRF integrates Region Purification to identify informative regions and leverage reliable and comprehensive regional cues for fine-grained alignment,
Cache Purification for constructing discriminative class-specific prototypes from purified regions for reliable calibration,
and temporal refreshing for improving long-term cache adaptability under streaming data.
% We introduce Multi-granularity Consistency Alignment (MCA), which enforces global–local consistency and refines semantic distributions via residual optimization to generate reliable pseudo-labels for multi-label objectives.
% \item We design Disentangled Region Prototype Caching (DRPC) and Temporal Cache Refreshing (TCR) to decouple regional prototypes from entangled global features and incorporate temporal sensitivity into cache updating, collectively achieving fine-grained representation learning and dynamic adaptability during prolonged adaptation.
\item Extensive experiments across five standard multi-label benchmarks and four backbones show that PuRF consistently outperforms the strongest state-of-the-art method, demonstrating its superior effectiveness.
\end{itemize}

%-------------------------------------------------------------------------

\section{Related Work}
\label{sec:related}

\noindent\textbf{Vision-Language Test-Time Adaptation.} 
% 可参考CVPR25 StatA
% 这个要怎么写啊！ 
% TTA是什么，最近VLM的TTA Boost了（开题报告里面的结合）
% prompt-based methods [cite] pioneer the use of ... 具体方案是怎么样的但是存在什么问题
% 后续发展起来的cache-based methods 效率更高而且精度也更好，de facto ...
% However, 这些方案即使高效有用，但是都集中在单标签
% ML-TTA首次将目光引入多标签的TTA，但是需要retrieval captions 和 反复重置更新, which 是比较耗时的. Our methods 首次探究cache-based的方法在 ML-TTA场景下应用潜力
% Unlike prior cache-based methods tailored for single-label classification, we extend cache-based TTA to the multi-label regime, addressing feature entanglement and prediction uncertainty caused by label co-occurrence and fine-grained semantic overlap.
% }
% Test-time adaptation (TTA) is developed for improving model generalization under distribution shifts without using any labeled data.
% The emergence of pre-trained VLMs has brought new momentum to TTA, enabling it to tackle more challenging open-world shifts and attracting considerable attention.
Test-time adaptation (TTA) improves model generalization under distribution shifts, and the rise of vision-language models (VLMs) further extends it to more challenging open-world shifts.
Existing TTA methods for VLMs can be broadly categorized into prompt-based and cache-based methods.
% prompt方法的缺点
% Prompt-based methods~\cite{tpt,difftpt,histpt,dart,ctpt}, pioneer this field by optimizing textual prompts during inference to reduce prediction uncertainty, but their efficiency is constrained by resetting the prompt parameters to initial state for each sample.
Prompt-based methods~\cite{tpt,histpt,dart,ctpt} pioneer this field by optimizing textual prompts via marginal entropy minimization, but suffer from limited efficiency due to per-sample prompt resetting and heavy backpropagation.
Cache-based methods~\cite{tda,dmn,cosmic,multicache} offer a more efficient alternative by constructing a dynamic cache of test features and refining predictions through retrieval-based similarity aggregation.
% have emerged as an efficient alternative, constructing a dynamic cache from test features to refine predictions through retrieval-based similarity aggregation.
% have recently gained attention as a prevailing approach for efficient and effective adaptation.
% By constructing a dynamic memory bank (or ``cache'') from historical test sample features, these approaches effectively model the target domain distribution on-the-fly and refine predictions via visual retrieval and similarity aggregation.
% TDA~\cite{tda} assists predictions via a positive and a negative cache; BoostAdapter~\cite{boostadapter} incorporates regional-level bootstrapping from each test sample; DPE~\cite{dpe} jointly optimizes cached visual and textual prototypes for better alignment; and ReTA~\cite{reta} further addresses both cache reliability and distribution calibration.
A line of work~\cite{tda,dmn,boostadapter} is training-free and focuses on the design of multiple caches, where different caches serve distinct roles to enhance adaptation.
Another line~\cite{dpe,reta} introduces lightweight prototype evolution through residual updates for better cross-modal alignment.
Despite advancements, most current methods mainly focus on single-label classification. 
Motivated by this gap, we extend the cache-based paradigm to multi-label adaptation by addressing challenges in purified regional cue selection and utilization, while improving discriminative class-wise cache calibration.

% \textcolor{red}{specifically tackling feature entanglement and fine-grained recognition challenges.}
% Unlike prior cache-based methods tailored for single-label classification, we extend cache-based TTA to the multi-label regime, addressing feature entanglement and prediction uncertainty caused by label co-occurrence and fine-grained semantic overlap.
% ------------------------------------------------------------------

\begin{figure*}[t!]
	
	\begin{minipage}[b]{1.0\linewidth}
		\centering
        \vspace{-0.05cm}
		\centerline{\includegraphics[width=11.9cm]{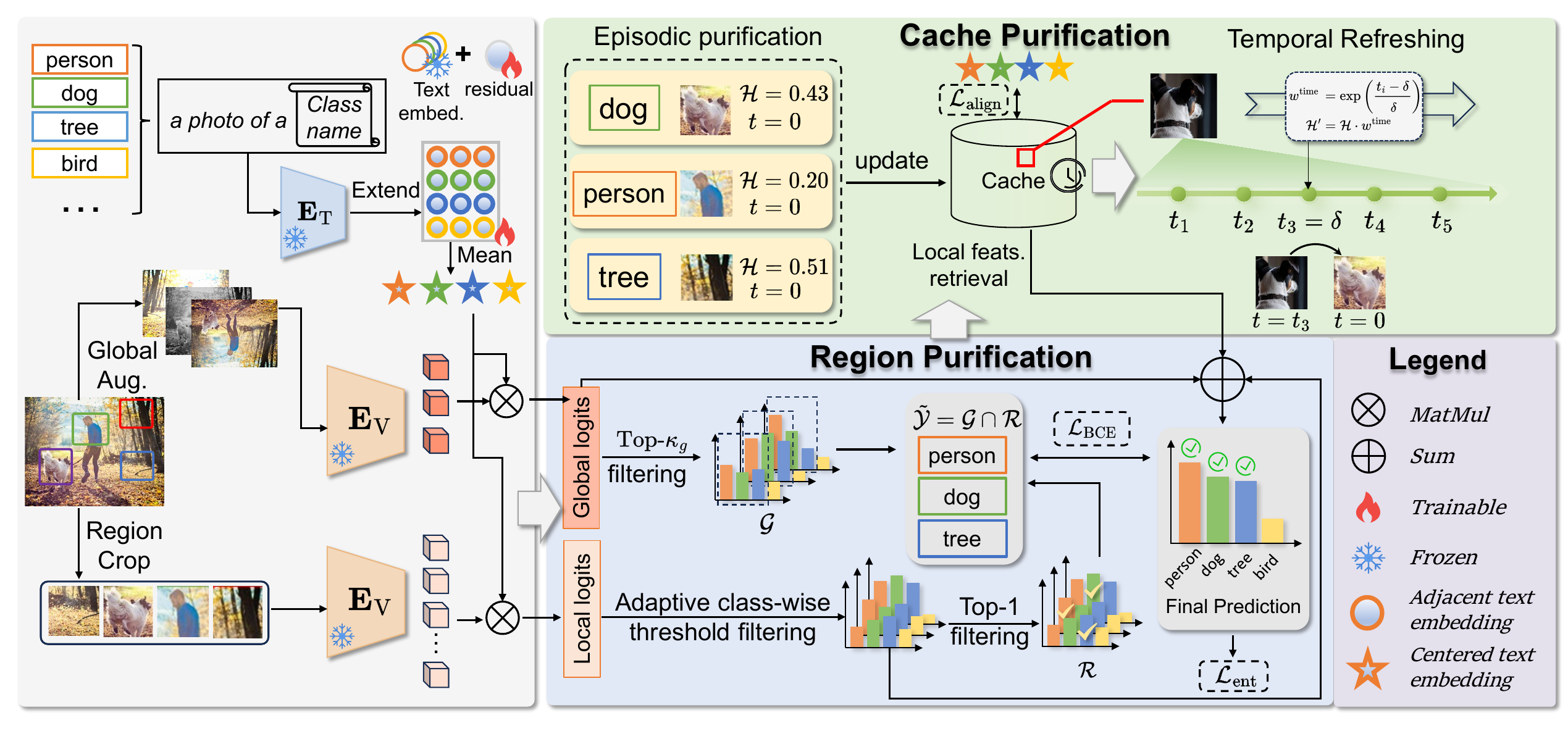}}
		%  \vspace{2.0cm}
	\end{minipage}
        % \vspace{-0.7cm}
        % 步骤是什么? 
        \vspace{-0.4cm}
	\caption{
    The overall framework of our proposed PuRF. 
    We first adaptively identify purified regions for prediction aggregation and then derive reliable pseudo supervision by enforcing global-local prediction consistency. 
    Guided by the pseudo labels, confident region-text pairs are stored as class-specific region prototypes for episodic cache purification, improving class-wise discriminability. 
    Temporal refreshing further purifies the cache by incorporating newly reliable samples while maintaining long-term adaptivity of cache.
    $\mathcal{H}$ denotes the prediction entropy.
    }
	\label{fig: pipeline}
	\vspace{-0.4cm}
\end{figure*}

\noindent\textbf{Multi-Label Recognition with VLMs.}
%目前有 Vision-Language Models for MLR
% VLM带来多标签学习的新范式（尤其在limited annotations下） → 大多使用prompt learning
% ③ 但这些方法仍假设有部分标注 →
% ⑥ 我们探索并扩展 fully 无标注场景下的挑战。
% \textcolor{red}{Traditional multi-label recognition explicitly models label dependencies using graph or transformer architectures, but relies on fully annotated data.
% With the rise of vision-language models (VLMs), label semantics can be expressed in natural language, enabling efficient multi-label learning under limited annotations.
% Recent works such as CLIP-MLL, DualCoOp, and MaPLe adapt CLIP through prompt or adapter tuning, achieving promising generalization.
% However, these approaches still assume access to labeled samples or training-time optimization, making them unsuitable for test-time scenarios.
% Multi-label test-time adaptation (ML-TTA) is a more realistic yet unexplored setting that requires unsupervised online adaptation under label co-occurrence and semantic overlap.
% Although cache-based TTA methods have shown strong adaptability and efficiency in single-label tasks, their potential in multi-label adaptation remains underexplored.
% We thus extend the cache-based paradigm to multi-label settings, enhancing adaptation through reliable caching and adaptive consistency alignment.}
% Traditional multi-label recognition relies on fully supervised learning and explicit modeling of label dependencies, which limits scalability in real-world scenarios. 
Multi-label recognition (MLR) serves as a fundamental task in computer vision, where the key challenges lie in modeling label dependencies~\cite{scpnet,HSPNet} and identifying discriminative properties for each label~\cite{discri1,discri2}.
Earlier methods rely solely on visual features and capture label correlations via graph structures~\cite{graph2,graph3}, recurrent networks~\cite{cnnrnn,rnn2}, or attention mechanisms~\cite{attention1}.
Recently, the emergence of VLMs has introduced a new paradigm for MLR, enabling strong zero-shot and open-vocabulary capabilities, particularly when combined with parameter-efficient tuning techniques such as prompt learning.
% VLMs offer a new paradigm by leveraging semantic-rich language prompts for zero-shot recognition with impressive generalization ability. Recent approaches combine VLMs with parameter-efficient fine-tuning (e.g. prompt tuning) to improve multi-label performance  even under limited supervision.
Methods like DualCoOp~\cite{dualcoop}, TaI-DPT~\cite{tai}, and PVP~\cite{pvp} introduce learnable prompts to capture label semantics, effectively modeling label co-occurrence with minimal additional parameters.
% adapt CLIP to better capture label semantics and co-occurrence. 
% 缺陷: 还是有labeled的问题
However, they still rely on annotated samples or offline fine-tuning on batches for stable adaptation.
In contrast, our work targets a fully unlabeled setting with online adaptation for each incoming sample, which remains challenging for precise multi-label recognition.
% 还有online的范式! instance-wise! 

% ------------------------------------------------------------------

\section{Methodology}
\label{sec:method}
% 需要说明我们的整个pipeline show在fig. 2
% In this section, we introduce ReD, a Region-Disentangled method for Multi-Label Test-Time Adaptation, as shown in Fig.~\ref{fig: pipeline}.
% 需要说明，哪一个模块对应哪一节! 
The overall pipeline of PuRF is shown in Fig.~\ref{fig: pipeline}.
PuRF enhances multi-label test-time adaptation through purification at the region and cache levels.
Sec.~\ref{sec:mca} proposes Region Purification (RP) via Multi-granularity Consistency, which selects informative regions using adaptive relative activation to provide comprehensive visual cues, 
and exploits global-local prediction consistency to derive reliable pseudo supervision for semantic residual optimization and fine-grained alignment.
Sec.~\ref{sec:drpc} presents Cache Purification (CP), comprising Episodic Purification (EP) that constructs region-aware class-specific prototypes from purified regions for stronger discriminability,
and Temporal Refreshing (TR) that further enhances long-term cache adaptivity.
% Region-aware Cache Purification (RCP) that builds class-specific cache prototypes from purified regions for stronger discriminability.
% Sec.~\ref{sec:tcr} further introduces the Temporal Cache Refreshing (TCR) strategy, which maintains cache adaptability under evolving test distributions.
Finally, Sec.~\ref{sec:loss} establishes a unified objective that enables joint optimization and feature alignment.

% It derives reliable supervision via multi-granularity consistency and progressively refines cached features into class-specific prototypes for stronger discriminability.
% Moreover, a temporal cache refreshing strategy further enhances cache adaptability under evolving distributions.
% Finally, a unified objective enables joint optimization and feature alignment with lightweight residual updates for better adaptation.
% which performs fine-grained and effective alignment between global and local branches to eliminate feature entanglement and enable precise multi-label recognition.
% ReD captures reliable supervision for optimization through multi-granularity consistency between global and local representations, while the cached features are progressively refined into class-specific prototypes to enhance discriminability. 
% Moreover, a temporal cache refreshing strategy is introduced to further improve flexibility and stability. Finally, under a unified objective, the framework achieves joint optimization and feature alignment, ensuring more robust and stable adaptation in multi-label scenarios.

\subsection{Preliminaries}

\noindent\textbf{Cache-based TTA for VLMs.}
Cache-based TTA methods~\cite{tda,dpe,boostadapter} build on the key-value design of \cite{tipadapter} that store low-entropy samples and continually update during inference. For image $x$ and a class prompt $\mathcal{T}^c$,
given the CLIP visual feature $\mathbf{F}_{\mathrm{V}} = \mathbf{E}_{\mathrm{V}}(x)$ and textual feature $\mathbf{F}_{\mathrm{T}}^{c} = \mathbf{E}_{\mathrm{T}}(\mathcal{T}^{c})$ extracted by its visual and text encoders $\mathbf{E}_{\mathrm{V}}$ and $\mathbf{E}_{\mathrm{T}}$. The prediction logits and probabilities are obtained from the cosine similarities $\langle \cdot , \cdot \rangle$:
% 要不然这个不要C了？ 
\begin{equation}
    \setlength{\abovedisplayskip}{4.5pt}
    \setlength{\belowdisplayskip}{5.5pt}
s_i^c=\left\langle\mathbf{F}_{\mathrm{V}}, \mathbf{F}_{\mathrm{T}}^c\right\rangle
% , \quad p_i=\operatorname{Softmax}\left(s_i\right)
\end{equation}
The cache $\mathcal{M} = \{\mathcal{M}^{c}\}_{c=1}^{\mathcal{C}}$ is composed of $\mathcal{C}$ class-wise subsets, each with a capacity of $L$ entries containing image features and their entropy values.
% Each class-wise cache is constructed as $\mathcal{M}^c$ consists of entries formed by image features and their self-entropy values, with a predefined capacity $L$.
% ----------------------------
%  each cache item is represented as:
% \begin{equation}
%     \left\{\mathbf{F}_{\text {cache }}, \mathbf{L}_p, H\left(\mathbf{F}_{\text {cache }} \mathbf{W}_{\scriptsize \text {CLIP }}^{\top}\right)\right\},
% \end{equation}
For test-time samples arriving on the fly, new items enter the corresponding cache slots based on their one-hot pseudo-labels. Once the cache reaches capacity, items with the highest entropy are replaced.
At inference, test feature $\mathbf{F}_{\mathrm{V}}$ retrieves relevant cache items,  and the prediction combines CLIP logits with cache logits:
% \begin{equation}
% \label{eq:cls_logits}
% \text{logits}_{\mathrm{cls}}(\mathbf{z}) = \mathbf{z}\mathbf{W}_\text{CLIP}^{\top} + \mathcal{A}(\mathbf{z}\mathbf{F}_{\text{cache}}^{\top})\mathbf{L}_p
% \end{equation}
\begin{equation}
    s_{\mathrm{TTA}}^{c}
    = 
    s_{i}^{c}
    + 
    s_{\text{cache}}^{c}
    =
    s_{i}^{c}
    + 
    \mathcal{A}\!\left(
    \langle
    \mathbf{F}_{\mathrm{V}} ,
    \mathbf{F}_{\text{cache}}^{c} \rangle
    \right)
    \mathbf{L}_{p}
    \label{eq:tta_score}
\end{equation}
where $\mathbf{F}_{\text{cache}}^{c}$ denotes cached prototypes of class $c$, and $\mathbf{L}_{p}$ is the one-hot pseudo-label from CLIP prediction.
$\mathcal{A}(x)=\alpha e^{-\beta(1-x)}$ is a modulating function with amplitude $\alpha$ and sharpness $\beta$.
Akin to \cite{dpe} and \cite{reta}, $\mathbf{F}_{\text{cache}}^{c}$ is obtained by averaging features in $\mathcal{M}^{c}$ for compact visual prototypes.

% Beyond the totally training-free cache-based TTA methods, several recent studies [cite] introduce lightweight residual learning with entropy-driven objectives on the output features, to further enhance adaptivity and adaptation performance without incurring substantial computational overhead.
% Since the residual learning operates directly on the encoded visual or textual prototype features, the gradients do not need to be back-propagated through the entire encoder.
% 除了我们的ReTA， 还需要再找一个工作才行！ 
% 记住一下，下面的要放在3,3最后！ 
% Beyond training-free cache-based TTA methods, 
Some cache-based methods~\cite{dpe,multicache,reta} introduce lightweight residual learning with entropy-driven objectives to further enhance adaptation without heavy computation. 
Recently, ReTA~\cite{reta} proposes adjacent textual embeddings that extend each class prototype into multiple semantic experts, with learnable residuals for distribution calibration, enabling more fine-grained contextual semantics.
% Following ReTA~\cite{reta}, we apply textual residual learning in our framework, formulated as:
Following this design, we implement textual residual learning as:
\begin{equation}\label{eq:textresidual}
    \setlength{\abovedisplayskip}{4.0pt}
    \setlength{\belowdisplayskip}{5.0pt}
\widehat{\mathbf{F}}_{\mathrm{T}}^{c} = \mathbf{F}_{\mathrm{T}}^{c} + \Delta\mathbf{F}_{\mathrm{T}}^{c}
\end{equation}
where $\Delta\mathbf{F}_{\mathrm{T}}^{c}$ denotes the learnable residual and the original prototype is frozen. 
We incorporate adjacent text embeddings as in~\cite{reta} to enhance semantic diversity, averaging them to form the textual prototype.
Additionally, we also progressively update the shared semantic embedding via running averaging over streaming test samples, following~\cite{reta,dpe}:
% \begin{equation}
%     \setlength{\abovedisplayskip}{2.0pt}
%     \setlength{\belowdisplayskip}{2.0pt}
%     \label{updating}
%     \widehat{\boldsymbol{t}}_m^c = \frac{(l-1)\widehat{\boldsymbol{t}}_m^c + \widehat{\boldsymbol{t}}_{m}^{c(\ast)}}{||(l-1)\widehat{\boldsymbol{t}}_m^c + \widehat{\boldsymbol{t}}_{m}^{c(\ast)}||}
% \end{equation}
\begin{equation}
\setlength{\abovedisplayskip}{2.0pt}
\setlength{\belowdisplayskip}{2.0pt}
\label{updating}
\widehat{\mathbf{F}}_{\mathrm{T}}^{c}
=
\operatorname{Norm}\!\left(
(l-1)\widehat{\mathbf{F}}_{\mathrm{T}}^{c}
+
\widehat{\mathbf{F}}_{\mathrm{T}}^{c(\ast)}
\right)
\end{equation}
where $\operatorname{Norm}(\mathbf{x})=\mathbf{x}/\|\mathbf{x}\|$ denotes $\ell_2$ normalization, and $l$ is the number of accumulated updates.
However, we do not adopt the entropy reweighting strategy, as it is tailored for single-label consistency.

% ----------------------------------------------------------------------------%
% \noindent\textbf{Semantic Residual Learning.}
% Recent studies in vision–language adaptation, such as DPE, BCE, and ReTA~\cite{ReTA}, have introduced the concept of \emph{semantic residual learning} to refine textual representations without disrupting the pre-trained knowledge of vision–language models.
% Instead of learning completely new prompts or textual prototypes, these methods learn a lightweight residual term that captures the semantic discrepancy between the pre-trained text features and the target-domain semantics:
% \begin{equation}
% \widehat{\mathbf{F}}_{\mathrm{T}}^{c} = \mathbf{F}_{\mathrm{T}}^{c} + \Delta\mathbf{F}_{\mathrm{T}}^{c},
% \end{equation}
% where $\mathbf{F}_{\mathrm{T}}^{c}$ denotes the frozen pre-trained textual feature of class $c$, and $\Delta\mathbf{F}_{\mathrm{T}}^{c}$ represents the learnable residual that encodes domain- or context-specific semantics.
% This residual formulation allows efficient adaptation while preserving the generalization ability of pre-trained vision–language models.

% 包括entropy计算，残差形式，以及讨论和别的工作的不同! 

% ----------------------------------------------------------------------------%

\subsection{Region Purification via Multi-granularity Consistency}~\label{sec:mca}
%%%%%%%%%%%%%%%%%%%%%%%% origin version %%%%%%%%%%%%%%%%%%%%%%%%%%%%%%%
% % 不，我觉得这里应该改成，更加claim 全局特征不足的问题! 而不是单纯地 global ... ? 
% % Conventional TTA methods directly minimize prediction entropy, which is suboptimal in multi-label settings as it suppresses potentially positive labels and overemphasizes dominant ones, hindering adaptation by ignoring label co-occurrence.
% % Hence, it is more suitable as an auxiliary constraint rather than the main objective.
% To adapt effectively to multi-label distributions, we employ the widely used binary cross-entropy (BCE) loss to independently identify label existence.
% Since ground-truth labels are unavailable at test time, BCE fully relies on pseudo-labels, whose reliability is crucial for adaptation.
% % This motivates our two-stage pseudo-label selection scheme based on global and local consistency, designed to provide more reliable supervision for test-time optimization.
% This motivates us to design a rigorous pseudo-label selection mechanism to provide more reliable supervision.
% To this end, we construct reliable pseudo-labels by leveraging consistency signals from both global and local perspectives.
%%%%%%%%%%%%%%%%%%%%%%%% fixed version %%%%%%%%%%%%%%%%%%%%%%%%%%%%%%%
Conventional TTA methods~\cite{tent,tda,dmn} typically rely on a global representation. However, in multi-label scenarios, such representations entangle co-occurring objects and are dominated by salient labels occupying large space~\cite{OFOL1,DCLIP}, making them insufficient for modeling fine-grained label-specific features.
An intuitive approach is to incorporate region-level cues via region cropping~\cite{beyond_object_proposal,cropping2,tagclip} or localization~\cite{WSOD,localization2}, which can isolate object-related evidence and benefit multi-label recognition.
% 这里先不强调supervision
However, region candidates often contain redundancy and noise, especially in unsupervised TTA settings. 
Naively aggregating all regions may dilute discriminative signals and introduce irrelevant context~\cite{RAM,Query-Based-Knowledge}, thereby limiting recognition performance.
This motivates us to design a region purification mechanism to select informative regions and facilitate fine-grained alignment to optimize semantic representations.
% Moreover, effective region selection and utilization are determinative for reliable adaptation, which motivates us to design a region purification mechanism to select informative regions and facilitate fine-grained semantic alignment based on purified regional evidence.
% Specifically, we selectively preserve informative regions to facilitate fine-grained semantic alignment.
% Specifically, We address this by introducing a region purification mechanism that selectively preserves informative regions and facilitates fine-grained semantic alignment.
% Specifically, we construct purified supervision signals by leveraging multi-granularity consistency from both global and local perspectives, guiding stable and discriminative adaptation.

% \textbf{Reliable Pseudo-labels Set Construction. }
\textbf{Adaptive Region Purification. }To support reliable regional evidence, we leverage both global and local cues.
At the global level, we apply holistic augmentations (\textit{e.g.}, flipping and color jitter) to form a multi-view set $\mathcal{V}$ following~\cite{boostadapter,tda}.
Each augmented view $x_i$ yields predictions $p(x_i)$, which are used to derive pseudo-label candidates to guide the purification and semantic alignment:
% \begin{equation}
% \mathcal{G}=\left\{k \mid k \in \bigcap_{v \in \mathcal{V}_f} \operatorname{Top}-\tau_g\left(p^{(v)}\right)\right\}
% \end{equation}
% \begin{equation}
%     \setlength{\abovedisplayskip}{5.0pt}
%     \setlength{\belowdisplayskip}{5.5pt}
% \scalebox{0.95}{$
%     \displaystyle
%     \mathcal{G}=\bigcap_{x_i \in \mathcal{V}} \mathrm{Top}\text{-}\kappa_g\left(p\left(x_i\right)\right).
%     \label{eq:globalcandidate}
%     $}
% \end{equation}
% \scalebox{0.95}{$
\begin{equation}
    \setlength{\abovedisplayskip}{3.7pt}
    \setlength{\belowdisplayskip}{4.5pt}
\scalebox{0.95}{$
\displaystyle
\mathcal{G}
=
\bigcap_{x_i \in \mathcal{V}}
\left\{
c \mid
c \in
\mathrm{Top}\text{-}\kappa_g
\left(p(x_i)\right)
\right\}.
    $}
\label{eq:globalcandidate}
\end{equation}
% This operation retains the classes that are consistently ranked within the top-$K_\mathrm{g}$ across multiple views.
% where $\kappa_g = \lfloor \rho_g \mathcal{C} \rfloor$ denotes the number of retained classes, and $\rho_g \in (0,1)$ is the retention ratio.
% where $\mathrm{Top}\text{-}\kappa_g(\cdot)$ returns the indices of the top-ranked $\kappa_g$ classes for initial candidate selection based on relative confidence, with $\kappa_g$ set as a fixed proportion of the total class number $\mathcal{C}$.
where $\mathrm{Top}\text{-}\kappa_g(\cdot)$ identifies the indices of the highest-confidence classes, with the threshold $\kappa_g$ defined as a fixed proportion of the total class count $\mathcal{C}$.
% selects the top-$\kappa_g$ classes based on relative confidence, with $\kappa_g$ set as a fixed proportion of the total class number $\mathcal{C}$.

% At the local level, 
% % given the global feature $\mathbf{F}_{\mathrm{V}} \in \mathbb{R}^{D}$, 
% we obtain a set of $Q$ local region features by performing random multi-scale cropping on the original image:
% \begin{equation}
% \scalebox{0.97}{$
%     \begin{aligned}
%     \mathcal{F}_{\mathrm{V}}^{\text{local}} 
%     &= \left\{\mathbf{f}_{\mathrm{V}}^{(1)}, \mathbf{f}_{\mathrm{V}}^{(2)}, \ldots, \mathbf{f}_{\mathrm{V}}^{(Q)}\right\}, \\
%     \mathbf{f}_{\mathrm{V}}^{(q)} 
%     &= \mathbf{E}_{\mathrm{V}}\!\left(\Phi_{\mathrm{crop}}^{(q)}(x)\right), \quad q = 1, 2, \ldots, Q
%     \end{aligned}
%     $}
% \end{equation}
At the local level, we extract $Q$ local region features 
{\small$\mathcal{F}_{\mathrm{V}}^{\text {local}}= \{\mathbf{f}_{\mathrm{V}}^{(1)}, \mathbf{f}_{\mathrm{V}}^{(2)}, \ldots, \mathbf{f}_{\mathrm{V}}^{(Q)}\}$ }
by performing stochastic scaling and cropping on the input image $x$ under predefined scale ranges, following~\cite{lazsl,beyond_object_proposal}.
We aim to identify regions that provide more isolated and less entangled label-specific evidence, which helps compensate for missing spatial cues and improve multi-label discrimination~\cite{Chen_Wang_Li_Lin_2018}, particularly for small or spatially localized objects~\cite{SRN}.
Intuitively, a purified region should exhibit clear dominance of one class over the others.
To quantify this dominance, we compute the relative confidence of each region via normalization across classes:
% \begin{equation}
% \scalebox{0.999}{$
% \hat{s}_q^{c} =
% \frac{
% \exp\!\left(
% \left\langle \mathbf{f}_{\mathrm{V}}^{(q)}, \mathbf{F}_{\mathrm{T}}^{c} \right\rangle
% \right)
% }
% {
% \sum_{j=1}^{\mathcal{C}}
% \exp\!\left(
% \left\langle \mathbf{f}_{\mathrm{V}}^{(q)}, \mathbf{F}_{\mathrm{T}}^{j} \right\rangle
% \right)
% }.
% $}
% \end{equation}
\begin{equation}
    \setlength{\abovedisplayskip}{4.0pt}
    \setlength{\belowdisplayskip}{5.0pt}
\hat{s}_q^{c} = \mathrm{softmax}_c \!\left(
\left\langle \mathbf{f}_{\mathrm{V}}^{(q)}, \mathbf{F}_{\mathrm{T}}^{c} \right\rangle
\right).
\end{equation}
We identify the dominant class for each region as 
$c^* = \arg\max_{c \in \mathcal{C}} \hat{s}_q^c$
and use the normalized score $\hat{s}_q^{c^*}$ as the purification confidence. 
Since confidence varies across classes due to class imbalance~\cite{classimbalance1}, a uniform global threshold becomes suboptimal for region selection.
% We therefore maintain an adaptive class-wise threshold $\mu_c$, progressively updated from running statistics of relative confidence across processed test samples:
% \begin{equation}
% \theta_c = \frac{1}{N_c} \sum_{i=1}^{N_c} \hat{s}_i^{c}.
% \end{equation}
% A region is retained only if $s_q^{c^*} \ge \theta_{c^*}$.
% We therefore maintain an adaptive class-wise statistic $\mu_c$ using running averages of batch-wise mean relative confidence:
% \begin{equation}
% \mu_c = \frac{1}{T}\sum_{t=1}^{T}\bar{s}^c_{(t)}, 
% \quad 
% \bar{s}^c_{(t)}=\frac{1}{N_t}\sum_{q=1}^{N_t}\hat{s}_{q}^{c}.
% \end{equation}
% A region is retained only if $\hat{s}_{q}^{c^*} \ge \mu_{c^*}$.
We therefore maintain an adaptive class-wise threshold $\mu_c$ via running averaging of incoming region confidence scores:
\begin{equation}
\scalebox{0.999}{$
\mu_c \leftarrow \mu_c + \frac{1}{T}\big(\bar{s}_c - \mu_c\big),
\quad
\bar{s}_c = \frac{1}{Q}\sum_{q=1}^{Q}\hat{s}_q^{c}.
$}
\end{equation}
where $T$ is the number of test steps. 
% Regions whose dominant-class confidence surpasses the class-wise adaptive threshold are retained, forming the purified region feature set:
Regions whose dominant-class confidence surpasses the class-wise adaptive threshold are retained, forming the purified region feature set that provides reliable class-specific evidence:
\begin{equation}\label{eq:purified_region_feats}
    \setlength{\abovedisplayskip}{4.0pt}
    \setlength{\belowdisplayskip}{5.0pt}
\scalebox{0.95}{$
\mathcal{F}_{\mathrm{V}}^{\text{pur}}
=
\left\{
\mathbf{f}_{\mathrm{V}}^{(q)}
\;\middle|\;
\hat{s}_q^{c^*} \ge \mu_{c^*},
\;
q\in[1,Q]
\right\}.
$}
\end{equation}
% \begin{equation}
%     \setlength{\abovedisplayskip}{4.5pt}
%     \setlength{\belowdisplayskip}{5.5pt}
% \scalebox{0.95}{$
% \mathbf{f}_{\mathrm{V}}^{(q)} = \mathbf{E}_{\mathrm{V}}\!\left(x^{(q)}\right), 
% \quad q = 1, 2, \ldots, Q
% $}
% \end{equation}
% % where $\Phi_{\text{crop}}^{(q)}(\cdot)$ denotes the $q$-th random crop operation applied to the image. 
% where $x^{(q)}$ denotes the $q$-th cropped region, which corresponds to a small localized area treated as a single-label region.
% Each cropped feature covers a relatively small localized area and is treated as a single-label region.
% By aggregating the top-$1$ predictions from purified regions, we obtain the region-level pseudo-label set:
% Region-level pseudo-labels are obtained by aggregating the top-$1$ predictions from purified regions:
Region-level pseudo-labels are obtained by aggregating the top-$1$ predictions from purified regions, serving as local counterparts to the global pseudo-labels:
% \begin{equation}
% y_l^{(i)}=\arg \max _{c \in \mathcal{C}}\left\langle\mathbf{f}_{\mathrm{V}}^{(i)}, \mathbf{F}_{\mathrm{T}}^c\right\rangle
% \end{equation}
% \begin{equation}\label{regionset}
%     \setlength{\abovedisplayskip}{4.0pt}
%     \setlength{\belowdisplayskip}{5.0pt}
%     \displaystyle
%     \resizebox{.6\linewidth}{!}{$
%     \displaystyle
%         \mathcal{R}=\left\{y^{(q)} \mid y^{(q)}=\arg \max _{c \in \mathcal{C}}\left\langle\mathbf{f}_{\mathrm{V}}^{(q)}, \mathbf{F}_{\mathrm{T}}^c\right\rangle,\;
%         q \in [1, Q]
%         \right\}.
% $}
% \end{equation}
\begin{equation}\label{regionset}
    \setlength{\abovedisplayskip}{4.0pt}
    \setlength{\belowdisplayskip}{5.0pt}
    \displaystyle
    \resizebox{.58\linewidth}{!}{$
    \displaystyle
        \mathcal{R}=\left\{c^{*} \mid c^{*}=\arg \max _{c \in \mathcal{C}}\left\langle\mathbf{f}_{\mathrm{V}}^{(q)}, \mathbf{F}_{\mathrm{T}}^c\right\rangle,\;
        \mathbf{f}_{\mathrm{V}}^{(q)}\in\mathcal{F}_{\mathrm{V}}^{\text{pur}}
        \right\}.
$}
\end{equation}
% each image is divided into a set of smaller regions $\mathcal{R}$, and independent predictions are made for each region.
% The final pseudo-label set is obtained by enforcing global-local consistency via set intersection:
% $\widetilde{\mathcal{Y}}=\mathcal{G} \cap \mathcal{R}$.
Thus, purified regions provide reliable local supervision for semantic optimization, while the final pseudo supervision is obtained via intersection:
$\widetilde{\mathcal{Y}}=\mathcal{G} \cap \mathcal{R}$. 
% We provide further theoretical analysis of this mechanism in Supplementary Sec. 3.
In Supplementary Sec. 3, we provide a theoretical analysis that formally demonstrates the effectiveness of the proposed region purification mechanism.

% We thus obtain the global and local candidate sets defined as:
% \begin{equation}\label{eq:interlabel}
%     \setlength{\abovedisplayskip}{4.5pt}
%     \setlength{\belowdisplayskip}{5.5pt}
% \widetilde{\mathcal{Y}}=\mathcal{G} \cap \mathcal{R}.
% \end{equation}

\textbf{Global-Local Aggregation and Alignment. }
We employ global-local aggregation to capture complementary information by extending the prediction in Eq.~\eqref{eq:tta_score} to balance global and local contributions:
% Specifically, we extend the cache-based prediction in Eq.~\eqref{eq:tta_score} to jointly balance global and local contributions across different granularities:
\begin{equation}\label{Eq:allscore}
    \setlength{\abovedisplayskip}{3.7pt}
    \setlength{\belowdisplayskip}{4.1pt}
\resizebox{.68\linewidth}{!}{$
    \begin{aligned}
        s^{c}_\text{TTA} =
        &\;
        \underbrace{
        \tfrac{1}{2}
        \left\langle \mathbf{F}_{\mathrm{V}}, \mathbf{F}_{\mathrm{T}}^{c} \right\rangle
        }_{\text{Global}}
        +
        \underbrace{
        \tfrac{1}{2}
        \max_{\mathbf{f}_{\mathrm{V}}^{(q)} \in \mathcal{F}_{\mathrm{V}}^{\text{pur}}}
        \left\langle
        \mathbf{f}_{\mathrm{V}}^{(q)},
        \mathbf{F}_{\mathrm{T}}^{c}
        \right\rangle
        }_{\text{Local}}
        +
        s_{\text{cache}}^c
    \end{aligned}
    $}
\end{equation}
% where $\mathrm{Top}\text{-}K_l(\cdot)$ denotes mean pooling over the top-$K_l$ most confident regions.
where $\max(\cdot)$ preserves the strongest class-specific regional activation as~\cite{dualcoop++,beyond_object_proposal} to suppress noise. 
This aggregation integrates purified regional evidence for more fine-grained recognition.
% Instead of aggregating all region candidates, we selectively retain informative and diverse regions to preserve complementary cues and enhance recognition performance. Detailed empirical and theoretical analyses of this design are provided in the Appendix due to space limitations.
% , and the probability for class $c$ is then given by $p^k=\sigma\left(s_k\right)$.
% With reliable pseudo-labels from the above global-local consistency, standard multi-label learning becomes feasible, allowing the use of the common BCE loss for supervision:
With reliable pseudo-labels derived above, we adopt the standard BCE loss for multi-label optimization:
\begin{equation}\label{bce}
\resizebox{.78\linewidth}{!}{$
\mathcal{L}_\text{BCE}=-\sum_{c=1}^\mathcal{C}\left[\tilde{y}^c \log (\sigma(s^{c}_\text{TTA})) + (1-\tilde{y}^c) \log (1-\sigma(s^{c}_\text{TTA}))\right]
$}
\end{equation}
where $\sigma(\cdot)$ denotes the Sigmoid function.
Overall, the proposed aggregation and alignment leverage complementary visual cues from purified regions for more precise adaptation, while enabling mutual refinement between region purification and semantic optimization. 
% Instead of aggregating all region candidates, we selectively retain informative and diverse regions 
% to preserve complementary cues and improve fine-grained recognition. 
% Further theoretical analyses of this mechanism are provided in the Appendix.

% This multi-granularity alignment provides reliable pseudo-supervision and effectively stabilizes optimization in the absence of labels.

% 为了实现高效的并且effective的adaptation, inspired by recent show strong performance 方法reta，我们构建了这个语义残差分布较准学习。 具体来说  xxx 残差公式 
% 不同的是，我们并未使用他们提出的consistency-aware 熵重加权模块，since 这个只针对单标签下的 xxx有效
% 还缺少预测分数计算的公式！ 

% ----------------------------------------------------------------------------%
\subsection{Cache Purification with Episodic and Temporal Strategies}~\label{sec:drpc} %需要对称一下
% While cache-based TTA methods demonstrate high efficiency and strong performance in single-label settings by exploiting historical prototypes for prediction calibration, they struggle in multi-label scenarios by generally treating global features as class-wise prototypes.
% Since a global feature encodes all concepts present in a multi-label image, it inevitably entangles co-occurring classes within the same embedding.
% This causes \textit{feature entanglement}~\cite{entangle1,entangle2}, where representations of different categories within same image become intertwined in the feature space, preventing model from accurately distinguishing label semantics.
% 我觉得这里应该提的是，As mentioned before, cache-based methods typically 使用全局特征进行 retrieval and storage, which causes severe feature entanglement~\cite{entangle1,entangle2}, since a global feature encodes all concepts that inevitably entangles co-occurring classes within the same embedding.
Cache-based methods typically rely on global features for retrieval and storage. 
In multi-label scenarios, these features are often dominated by salient classes and lack label purity, leading to cache contamination and degraded class-wise calibration. 
% 拔高一点
To alleviate this issue, we perform cache purification at two complementary timescales: \textit{short-term} episodic purification and \textit{long-term} temporal refreshing.
The former ensures reliable cache entries for each incoming sample, while the latter enhances adaptation over time by refreshing outdated cache entries.

\textbf{Episodic Purification. }
For each incoming test sample (episode), we obtain its purified regions as defined in Eq.~\eqref{eq:purified_region_feats}. 
Given the reliable pseudo-label set $\widetilde{\mathcal{Y}}$, we select the corresponding regions to form region-label cache entries. 
% leading to severe feature entanglement~\cite{entangle1,entangle2} as co-occurring classes are blended within the same embedding.
% Such entanglement further contaminates the cache, hindering the progressive modeling for class-specific representation and compromising the exploitation of historical information.
% To mitigate this issue, we replace the image-level cache representation by reliable regions and associate them with corresponding labels, forming region-label pairs that serve as cache candidates to preserve class-wise discriminability.
% To alleviate this issue, we perform episodic purification for each incoming test sample, replacing the image-level cache candidate with the most reliable regions.
% Each region is associated with its corresponding label, forming region–label pairs that serve as purified cache candidates to preserve class-wise discriminability.
% we replace the image-level cache representation with purified regions and associate them with corresponding labels, forming region–label pairs as purified cache candidates that preserve class-wise discriminability.
% Specifically, we identify the most confident region corresponding to each pseudo-label, since each label in the reliable pseudo-label set $\widetilde{\mathcal{Y}}$ has at least one corresponding region obtained through intersection filtering.
% For each $\widetilde{y}_i \in \widetilde{\mathcal{Y}}$, 
% we select the corresponding region from the local feature set $\mathcal{F}_{\mathrm{V}}^{\text {local}}$ with the lowest prediction entropy as anchor:
Specifically, for each label $\widetilde{y}_i \in \widetilde{\mathcal{Y}}$, we select the most confident region in $\mathcal{F}_{\mathrm{V}}^{\text{pur}}$ as anchor:
\begin{equation}
% \resizebox{.6\linewidth}{!}{$
    \setlength{\abovedisplayskip}{4.0pt}
    \setlength{\belowdisplayskip}{5.0pt}
    \displaystyle
    r^{\star} =
    % \arg\min_{r \in \mathcal{\mathcal{F}_{\mathrm{V}}^{\text {local}}}}
    \arg\min_{\mathbf{f}_{\mathrm{V}}^{(q)} \in \mathcal{F}_{\mathrm{V}}^{\text{pur}}}
    \mathcal{H}\left(
    \left\langle
    \mathbf{f}_{\mathrm{V}}^{(q)},
    \mathbf{F}_{\mathrm{T}}^{\widetilde{y}_i}
    \right\rangle
    \right)
    % $}
    % \quad \widetilde{y}_i \in \widetilde{\mathcal{Y}}
\end{equation}
where region $r^{\star}$ denotes the anchor region index for pseudo-label $\widetilde{y}_i$.
Each anchor feature $\mathbf{f}_{\mathrm{V}}^{({r}^{\star})}$ is paired with its entropy to form region-entropy cache candidates.
% This simple but effective design alleviates feature entanglement in the cache, as empirically demonstrated in Fig.~\ref{fig: tsne}. 
This design yields cleaner cache representations and improves class-wise discriminability, as empirically demonstrated in Fig.~\ref{fig: tsne}. 
% The cache score is computed over the same purified region set to maintain granularity consistency:
The cache score is computed on the same purified region set as Eq.~\eqref{Eq:allscore}:
\begin{equation}
    \setlength{\abovedisplayskip}{4.0pt}
    \setlength{\belowdisplayskip}{5.0pt}
    \resizebox{.45\linewidth}{!}{$
    \displaystyle
    s_{\text{cache}}^{c}
    =
    \mathcal{A}\!\left(
    \max_{\mathbf{f}_{\mathrm{V}}^{(q)} \in \mathcal{F}_{\mathrm{V}}^{\text{pur}}}
    \left\langle
    \mathbf{f}_{\mathrm{V}}^{(q)} ,
    \mathbf{F}_{\text{cache}}^{c}
    \right\rangle
    \right)
    \mathbf{L}_{p}.
    \label{eq:newcache}
    $}
\end{equation}

Unlike typical entropy-based selection that stores a single cache entry per sample~\cite{tda,dpe}, our episodic purification exploits purified regions to form region-label pairs, enabling multiple class-specific cache entries within each episode.
Cache logits are computed through region-level max aggregation over the purified regions, improving prediction robustness and class-wise discriminability.

% Thus, for each test sample, every class in the reliable pseudo-label set $\widetilde{\mathcal{Y}}$ corresponds to one regional anchor, forming a local feature set:
% \begin{equation}
% { \mathbf{f}*{r_k^{\star}} }*{k \in \widetilde{\mathcal{Y}}}
% \end{equation}
% When writing to the cache, each entry contains not only the regional feature but also its associated label and uncertainty information, represented as:
% \begin{equation}
% { \mathbf{f}*{r_k^{\star}}, k, H(\mathbf{f}*{r_k^{\star}} \mathbf{W}_{\text{CLIP}}^{\top}) }, \quad k \in \widetilde{\mathcal{Y}}
% \end{equation}

% Finally, the cache for each class $B_k$ consists of multiple regional entries from different images, which can be aggregated via mean or weighted pooling to form the regional prototype for that class, providing fine-grained and disentangled representations for subsequent matching and inference:

% \begin{equation}
% \mu_k^{\text{reg}} = \frac{1}{|B_k|} \sum_{(\mathbf{f}_j, k) \in B_k} \mathbf{f}_j.
% \end{equation}

% ----------------------------------------------------------------------------%

\begin{figure}[!t]
\centering
    % \vspace{-0.3cm}
    \centering
	\begin{minipage}[b]{1.0\linewidth}
		\centering
        % \vspace{-0.18cm}
		\centerline{\includegraphics[width=10.7cm]{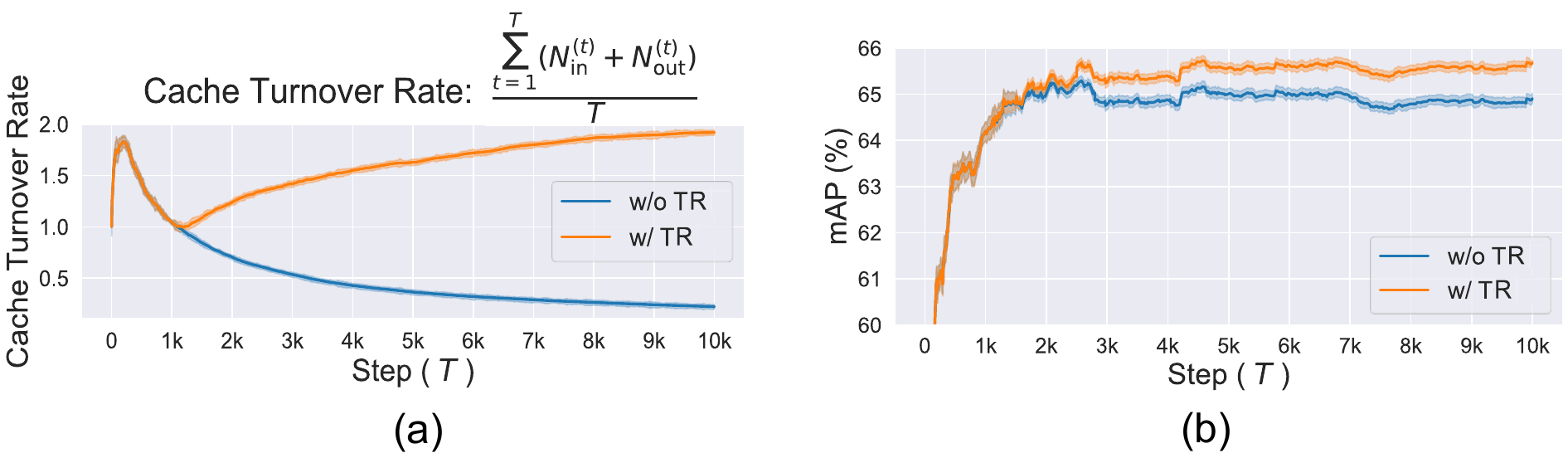}}
		%  \vspace{2.0cm}
	\end{minipage}
    \vspace{-0.45cm}
\caption{Empirical studies of cache dynamics on COCO2014. % 写定义，然后分析写在原文！ 
(a): The cache turnover rate measures the frequency of cache updates, where $N_{\mathrm{in}}^{(t)}$ and $N_{\mathrm{out}}^{(t)}$ denote the numbers of samples inserted and removed at adaptation step $t$, respectively.
(b): The evolution of model performance during adaptation.
% Cache turnover ratio and mAP evolution.
}
    \label{fig: 3}
    \vspace{-0.4cm}
\end{figure}

% \subsection{Temporal Cache Refreshing}~\label{sec:tcr}
\textbf{Temporal Refreshing. }As mentioned above, each multi-label image could produce multiple cache pairs, which accelerates cache saturation under limited capacity compared with single-label images that typically provide only one candidate.
As adaptation proceeds, the entropy-based admission threshold becomes increasingly stringent, making it difficult for later samples to enter the cache.
Consequently, the cache is rarely updated in the later stages and may eventually solidify into a fixed state.
We refer to this phenomenon as \textit{cache saturation}, where overly strict admission suppresses the inclusion of later samples, causing the cache to be dominated by early entries and gradually lose temporal dynamics --- its ability to update with samples appearing later in the data stream.
As shown in Fig.~\ref{fig: 3}, the blue curve denotes the vanilla cache without any adjustment.
Its turnover rate first increases and then declines, while the mAP curve plateaus later, indicating that the cache becomes progressively less adaptive over time.
As the cache becomes nearly fixed, it struggles to incorporate later samples, thereby overlooking their contribution.
Meanwhile, the progressive prototype update in Eq.~\eqref{updating} relies on accumulating semantic evidence over time.
When later samples fail to enter the cache, it cannot benefit from the progressively refined semantic prototypes for reliable entry selection, ultimately limiting long-term adaptation.

To maintain cache purification and enhance long-term adaptability, we design a Temporal Refreshing mechanism with a time-decay strategy to adjust the priority of cached entries, 
% preventing early samples from dominating the cache and promoting the adaptive integration of recent ones. 
mitigating early-stage bias that causes early samples to dominate the cache and enabling the integration of later ones.
To be specific, we additionally record for each cached item its cumulative time in the cache, denoted as $t_i$, and each class-wise cache $\mathcal{M}^c$ consists of $L$ feature entries represented as:
% Accordingly, each class-wise cache $\mathcal{M}^c$ consists of $L$ feature entries represented as:
\begin{equation}\label{memoryeq}
\mathcal{M}^c=\left\{\left({\mathbf{F}}_{\text {cache }}, \mathcal{H}\left({\langle\mathbf{F}}_{\text {cache }} ,{\mathbf{F}}_{\mathrm{T}}^c\rangle\right), t_i\right)\right\}_{i=1}^L 
\end{equation}
 % To address this issue, we record the storage time step of each sample and design a time-based priority decay strategy, which exponentially penalizes the entropy of long-stored entries to lower their admission priority and promote replacement with new samples.
 
Inspired by the exponential decay strategy in learning rate scheduling~\cite{sgdr,li2019exponential},
we formulate a similar temporal decay strategy to gradually downweight the priority of long-retained cache entries by penalizing their entropy, mitigating early-sample dominance.
% reduce dominance of earlier ones and facilitate replacement by new samples.
% Let $\delta$ denote the decay onset, after which temporal penalization begins and $\kappa$ represent the decay scale, with the temporal weight and the updated entropy defined as:
Let $\delta$ denote a temporal decay constant that controls the retention time, beyond which the penalization progressively increases.
The temporal weight and updated entropy are defined as:
\begin{equation}\label{eq:tcr}
    \setlength{\abovedisplayskip}{3.5pt}
    \setlength{\belowdisplayskip}{4.3pt}
% \scalebox{0.98}{$
w_i^{\text {time }}=\exp \left(\frac{t_i-\delta}{\delta}\right), \quad \mathcal{H}_i^{\prime}=\mathcal{H}_i \cdot w_i^{\text {time }}.
% $}
\end{equation}
The updated entropy $\mathcal{H}_i^{\prime}$ replaces the original one in Eq.~\eqref{memoryeq} to penalize long-retained entries.
% Since the cache prioritizes low-entropy samples, 
When $t_i > \delta$, the weight $w_i^{\text{time}} > 1$ amplifies their entropy, lowering priority and encouraging replacement by newer samples.
This refreshing strategy enables the cache to evolve with streaming data, allowing later samples to leverage refined semantic representations to progressively improve class-specific representations, thereby enhancing overall adaptation (see Fig.~\ref{fig: 3}).

% and enhance adaptation (see Fig.~\ref{fig: 3}).

% the temporal adaptivity of the cache is enhanced, enabling effective alignment with evolving data distributions (see Fig.~\ref{fig: 3}).

\subsection{Overall Objectives}~\label{sec:loss}
% 以往一般TTA的优化表现不好
Our optimization goal is to achieve accurate recognition in multi-label test-time adaptation, which is challenging for conventional TTA methods that mainly rely on vanilla entropy minimization. 
% Such optimization often over-suppresses uncertain predictions, leading to biased adaptation and degraded recognition performance.
Based on the reliable pseudo-labels obtained from global-local consistency (Sec.~\ref{sec:mca}), 
% we formulate the adaptation as a standard multi-label learning problem using binary cross-entropy in Eq.xx, which offers a more stable supervision signal than using entropy alone.
we incorporate the binary cross-entropy loss in Eq.~\eqref{bce} to provide additional supervision and stabilize optimization under distribution shifts.
Following~\cite{reta,dpe}, we adopt the standard entropy loss and an alignment loss as auxiliary objectives to optimize textual residuals and calibrate predictions.
Given test sample $x$ and its augmented set $\mathcal{V}=\left\{x_1, \ldots, x_{N-1}\right\}$, the marginal entropy loss can be written as:
\begin{equation}
    \setlength{\abovedisplayskip}{4.0pt}
    \setlength{\belowdisplayskip}{4.5pt}
\scalebox{0.96}{$
    \displaystyle
    \mathcal{L}_{\text{ent}}\left(x\right)=\frac{1}{\left|\mathcal{V}_f\right|} \sum_{x_i \in \mathcal{V}_f} \mathcal{H}\left(p\left(x_i\right)\right)  
    % \mathcal{H}(p)=-\sum_{c=1}^\mathcal{C} p_c \log p_c,
    $}
\end{equation}
where $\mathcal{H}(p)=-\sum_{c=1}^\mathcal{C} p^c \log p^c$ is the prediction entropy, and $\mathcal{V}_f\subset\mathcal{V} \cup \{x\}$ denotes the subset of the most confident augmented views (typically the top 10\%).
% Specifically, the entropy loss $\mathcal{L}_\text{ent}$ reducing predictive uncertainty, while t
Following~\cite{dpe,reta}, we introduce an alignment loss to enforce cross-modal consistency between cached visual prototypes and text, formulated as:
% The alignment loss $\mathcal{L}_\text{align}$ enforces cross-modal consistency by aligning the adaptive textual prototypes with the cached visual prototypes in the InfoNCE~\cite{infonce} form, formulated as:
% 这个公式再修改一下typo!  这怎么修改啊？ 
% \begin{equation}
% \resizebox{.72\linewidth}{!}{$
%     \mathcal{L}_{\text{align}}
%     = -\log
%     \frac{
%         \exp\!\left(\left(\widehat{\mathbf{F}}_{\mathrm{T}}^{c}\right)^{\!\top}\!\mathbf{F}_{\text{cache}}^{c}\right)
%     }{
%         \sum_{j=1}^{\mathcal{C}}
%         \exp\!\left(\left(\widehat{\mathbf{F}}_{\mathrm{T}}^{c}\right)^{\!\top}\!\mathbf{F}_{\text{cache}}^{j}\right)
%     }
%     $}
% \end{equation}
\begin{equation}
    \setlength{\abovedisplayskip}{4.0pt}
    \setlength{\belowdisplayskip}{4.5pt}
\resizebox{.65\linewidth}{!}{$
    \mathcal{L}_{\text{align}}
    = 
    \sum_{c=1}^{\mathcal{C}}
    \left[
        -\log
        \frac{
            \exp\!\left(\left(\widehat{\mathbf{F}}_{\mathrm{T}}^{c}\right)^{\!\top}\!\mathbf{F}_{\text{cache}}^{c}\right)
        }{
            \sum_{j=1}^{\mathcal{C}}
            \exp\!\left(\left(\widehat{\mathbf{F}}_{\mathrm{T}}^{c}\right)^{\!\top}\!\mathbf{F}_{\text{cache}}^{j}\right)
        }
    \right]
    $}
\end{equation}
where $\widehat{\mathbf{F}}_{\mathrm{T}}^{c}$ and $\mathbf{F}_{\text{cache}}^{c}$ denote the mean prototypes of the textual and visual prototypes, respectively.
In summary, our final loss function is defined as:
\begin{equation}\label{all_loss}
    \setlength{\abovedisplayskip}{4.0pt}
    \setlength{\belowdisplayskip}{4.5pt}
\mathcal{L}^* = \mathcal{L}_\text{ent}  + \lambda_{1}\mathcal{L}_\text{BCE} + \lambda_{2}\mathcal{L}_\text{align}
\end{equation}
where $\lambda_{1}$ and $\lambda_{2}$ are trade-off coefficients balancing contributions of pseudo-label supervision and cross-modal alignment.
The integration of multi-label supervision with test-time regularization via residual updates enables PuRF to capture label co-occurrence knowledge and mitigate test-time uncertainty.

%%%%%%%%%%%%%%%%%% TABLE 1 %%%%%%%%%%%%%%%%%%%%%%%%

% 导言区需要：
% \usepackage{makecell}
% \usepackage{multirow}

\section{Experiments}

\subsection{Experimental Settings}
\textbf{Datasets. } Following~\cite{mltta}, we evaluate on five widely used multi-label datasets: VOC~\cite{voc} (versions 2007 and 2012), MS-COCO~\cite{mscoco} (versions 2014 and 2017), and NUS-WIDE~\cite{nuswide}. 
For the VOC datasets, the 2007 and 2012 versions contain 20 object categories with 4.9K and 5.8K test images, respectively.
MS-COCO covers 80 object categories, and we use the validation sets of COCO2014/COCO2017 containing 40K/5K testing images.
For NUS-WIDE, we evaluate on 81 manually annotated categories covering broader concepts, with 83K test images.
% All evaluation splits and settings follow those used in ~\cite{mltta} to ensure a fair comparison.

\begin{table}[t!]
    \centering
    \caption{Comparison with state-of-the-art methods under different visual encoder architectures \textit{w.r.t.} mAP (\%). Bold indicates best results.}
    \vspace{-0.1cm}
    \label{tab: 1}
    % \newcolumntype{C}{>{\centering\arraybackslash}X}
    \resizebox{\columnwidth}{!}{%
    \begin{tabular}{@{}l|cccccc|cccccc@{}}
        % ================= 第一部分：ResNet 系列 =================
        \toprule[1.5pt] % 顶端加粗线
        & \multicolumn{6}{c|}{\textbf{ResNet-50}} & \multicolumn{6}{c}{\textbf{ResNet-101}} \\
        Methods & VOC07 & VOC12 & COCO14 & COCO17 & NUS & \textbf{Avg.} & VOC07 & VOC12 & COCO14 & COCO17 & NUS & \textbf{Avg.} \\
        \midrule
        CLIP~\cite{clip}\textsubscript{\fontsize{6}{6}\selectfont ICML'21} & 75.91 & 74.25 & 47.53 & 47.32 & 41.53 & \cellcolor{avgpurple}57.31 & 76.72 & 74.21 & 48.83 & 48.15 & 41.93 & \cellcolor{avgpurple}57.97 \\
        \midrule
        TPT~\cite{tpt}\textsubscript{NeurIPS'22} & 75.54 & 73.92 & 48.52 & 48.51 & 41.97 & \cellcolor{avgpurple}57.69 & 74.82 & 73.39 & 49.71 & 48.89 & 43.10 & \cellcolor{avgpurple}57.98 \\
        DiffTPT~\cite{difftpt}\textsubscript{CVPR'23} & 75.89 & 74.13 & 48.56 & 48.67 & 41.33 & \cellcolor{avgpurple}57.72 & 74.98 & 74.31 & 49.45 & 49.19 & 42.93 & \cellcolor{avgpurple}58.17 \\
        RLCF~\cite{rlcf}\textsubscript{ICLR'24} & 65.75 & 64.73 & 36.87 & 36.73 & 29.83 & \cellcolor{avgpurple}46.78 & 71.21 & 69.63 & 40.53 & 39.79 & 31.77 & \cellcolor{avgpurple}50.59 \\
        ML-TTA~\cite{mltta}\textsubscript{ICLR'25} & 78.62 & 76.63 & 51.58 & 51.39 & 42.53 & \cellcolor{avgpurple}60.15 & 78.72 & 78.13 & 52.92 & 52.24 & 43.62 & \cellcolor{avgpurple}61.13 \\
        \midrule
        TDA~\cite{tda}\textsubscript{CVPR'24} & 76.64 & 75.12 & 48.91 & 49.11 & 42.34 & \cellcolor{avgpurple}58.42 & 78.12 & 77.13 & 50.19 & 49.78 & 43.13 & \cellcolor{avgpurple}59.67 \\
        DMN~\cite{dmn}\textsubscript{CVPR'24} & 74.87 & 74.13 & 44.54 & 44.18 & 41.32 & \cellcolor{avgpurple}55.81 & 76.82 & 75.32 & 46.28 & 45.44 & 42.71 & \cellcolor{avgpurple}57.31 \\
        DPE~\cite{dpe}\textsubscript{NeurIPS'24} & 82.51 & 80.64 & 54.38 & 54.74 & 45.96 & \cellcolor{avgpurple}63.65 & 82.55 & 81.21 & 54.92 & 54.53 & 46.64 & \cellcolor{avgpurple}63.97 \\
        BoostAdapter~\cite{boostadapter}\textsubscript{NeurIPS'24} & 79.53 & 78.00 & 52.07 & 52.92 & 44.20 & \cellcolor{avgpurple}61.34 & 81.13 & 80.06 & 53.90 & 54.33 & 45.47 & \cellcolor{avgpurple}62.98 \\
        ReTA~\cite{reta}\textsubscript{ACM MM'25} & 83.56 & 81.98 & 55.09 & 55.40 & 46.90 & \cellcolor{avgpurple}64.59 & 83.69 & 82.40 & 56.71 & 56.39 & 47.02 & \cellcolor{avgpurple}65.24 \\
        \cellcolor{avgblue}\textbf{PuRF (Ours)} & \cellcolor{avgblue}\textbf{86.35} & \cellcolor{avgblue}\textbf{84.38} & \cellcolor{avgblue}\textbf{61.47} & \cellcolor{avgblue}\textbf{61.56} & \cellcolor{avgblue}\textbf{47.57} & \cellcolor{avgpurple}\textbf{68.27} & \cellcolor{avgblue}\textbf{86.77} & \cellcolor{avgblue}\textbf{85.43} & \cellcolor{avgblue}\textbf{62.60} & \cellcolor{avgblue}\textbf{61.93} & \cellcolor{avgblue}\textbf{47.91} & \cellcolor{avgpurple}\textbf{68.93} \\
        
        \specialrule{0.8pt}{3pt}{3pt} % 中间拼接线，厚度0.8pt，上下留白各3pt
        
        % ================= 第二部分：ViT 系列 =================
        & \multicolumn{6}{c|}{\textbf{ViT-B/16}} & \multicolumn{6}{c}{\textbf{ViT-B/32}} \\
        Methods & VOC07 & VOC12 & COCO14 & COCO17 & NUS & \textbf{Avg.} & VOC07 & VOC12 & COCO14 & COCO17 & NUS & \textbf{Avg.} \\
        \midrule
        CLIP~\cite{clip}\textsubscript{ICML'21} & 79.58 & 79.25 & 54.42 & 54.13 & 45.65 & \cellcolor{avgpurple}62.61 & 77.18 & 76.85 & 50.31 & 50.15 & 42.90 & \cellcolor{avgpurple}59.48 \\
        \midrule
        TPT~\cite{tpt}\textsubscript{NeurIPS'22} & 77.54 & 77.39 & 53.32 & 54.20 & 46.15 & \cellcolor{avgpurple}61.72 & 74.21 & 71.93 & 48.12 & 48.63 & 43.63 & \cellcolor{avgpurple}57.30 \\
        DiffTPT~\cite{difftpt}\textsubscript{CVPR'23} & 77.93 & 77.24 & 53.91 & 54.15 & 46.13 & \cellcolor{avgpurple}61.87 & 74.50 & 72.98 & 48.73 & 49.19 & 43.42 & \cellcolor{avgpurple}57.76 \\
        RLCF~\cite{rlcf}\textsubscript{ICLR'24} & 79.29 & 79.26 & 54.21 & 54.43 & 43.18 & \cellcolor{avgpurple}62.07 & 77.12 & 76.83 & 50.28 & 49.59 & 43.29 & \cellcolor{avgpurple}59.42 \\
        ML-TTA~\cite{mltta}\textsubscript{ICLR'25} & 81.28 & 81.13 & 57.52 & 57.49 & 46.55 & \cellcolor{avgpurple}64.80 & 78.70 & 77.97 & 52.83 & 52.99 & 44.12 & \cellcolor{avgpurple}61.32 \\
        \midrule
        TDA~\cite{tda}\textsubscript{CVPR'24} & 80.12 & 79.92 & 55.21 & 55.46 & 46.72 & \cellcolor{avgpurple}63.49 & 77.62 & 77.12 & 51.23 & 51.49 & 44.13 & \cellcolor{avgpurple}60.32 \\
        DMN~\cite{dmn}\textsubscript{CVPR'24} & 79.83 & 79.67 & 52.52 & 52.37 & 46.27 & \cellcolor{avgpurple}62.13 & 77.42 & 76.60 & 49.32 & 48.13 & 43.41 & \cellcolor{avgpurple}58.98 \\
        DPE~\cite{dpe}\textsubscript{NeurIPS'24} & 84.10 & 83.62 & 59.21 & 59.28 & 48.90 & \cellcolor{avgpurple}67.02 & 82.38 & 81.74 & 55.79 & 55.32 & 47.11 & \cellcolor{avgpurple}64.47 \\
        BoostAdapter~\cite{boostadapter}\textsubscript{NeurIPS'24} & 83.01 & 82.56 & 58.07 & 58.64 & 47.76 & \cellcolor{avgpurple}66.01 & 80.82 & 80.03 & 54.49 & 54.75 & 46.14 & \cellcolor{avgpurple}63.25 \\
        ReTA~\cite{reta}\textsubscript{ACM MM'25} & 85.09 & 84.49 & 60.37 & 60.30 & 49.20 & \cellcolor{avgpurple}67.89 & 83.99 & 83.51 & 57.02 & 56.58 & 48.05 & \cellcolor{avgpurple}65.83 \\
        \cellcolor{avgblue}\textbf{PuRF (Ours)} & \cellcolor{avgblue}\textbf{88.18} & \cellcolor{avgblue}\textbf{87.12} & \cellcolor{avgblue}\textbf{66.05} & \cellcolor{avgblue}\textbf{65.33} & \cellcolor{avgblue}\textbf{50.72} & \cellcolor{avgpurple}\textbf{71.48} & \cellcolor{avgblue}\textbf{87.32} & \cellcolor{avgblue}\textbf{86.04} & \cellcolor{avgblue}\textbf{63.42} & \cellcolor{avgblue}\textbf{63.02} & \cellcolor{avgblue}\textbf{49.61} & \cellcolor{avgpurple}\textbf{69.88} \\
        \bottomrule[1.5pt] % 底端加粗线
    \end{tabular}}
    \vspace{-0.3cm}
\end{table}

\begin{table}[!t]
  \centering
    \caption{Comparison with state-of-the-art methods under different prompt initializations on ViT-B/16 \textit{w.r.t.} mAP (\%). Bold indicates best results.}
    \vspace{-0.1cm}
  \resizebox{0.89\columnwidth}{!}{%
    \begin{tabular}{l|ccccc|c}
    \toprule
    Methods & VOC2007 & VOC2012 & COCO2014 & COCO2017 & NUS-WIDE & Avg. \\
    \midrule
    \rowcolor{gray!10}
    \multicolumn{7}{c}{\textbf{\textit{CoOp}}} \\
    \midrule
    CoOp~\cite{coop}\textsubscript{\scriptsize IJCV'22}  & 79.14 & 77.85 & 56.12 & 56.35 & 46.74 & \cellcolor{avgpurple}63.24 \\
    \midrule
    TPT~\cite{tpt}\textsubscript{\scriptsize NeurIPS'22}   & 79.72 & 77.85 & 55.35 & 55.23 & 47.27 & \cellcolor{avgpurple}63.08 \\
    DiffTPT~\cite{difftpt}\textsubscript{\scriptsize CVPR'23} & 79.86 & 77.61 & 55.30 & 55.47 & 47.13 & \cellcolor{avgpurple}63.07 \\
    RLCF~\cite{rlcf}\textsubscript{\scriptsize ICLR'24}  & 80.15 & 78.24 & 56.72 & 56.18 & 47.62 & \cellcolor{avgpurple}63.78 \\
    ML-TTA~\cite{mltta}\textsubscript{\scriptsize ICLR'25} & 83.17 & 81.36 & 59.68 & 59.33 & 48.12 & \cellcolor{avgpurple}66.33 \\
    \midrule
    TDA~\cite{tda}\textsubscript{\scriptsize CVPR'24}   & 80.20 & 78.58 & 56.93 & 57.15 & 47.82 & \cellcolor{avgpurple}64.13 \\
    DPE~\cite{dpe}\textsubscript{\scriptsize NeurIPS'24}   &  84.97     & 83.86      & 61.39      & 61.04      & 48.49      & \cellcolor{avgpurple}67.95   \\
    BoostAdapter~\cite{boostadapter}\textsubscript{\scriptsize NeurIPS'24} & 82.32      & 81.71      & 58.77      & 58.02      &  47.56     & \cellcolor{avgpurple}65.68  \\
    ReTA~\cite{reta}\textsubscript{\scriptsize ACM MM'25}  & 85.23 & 84.15 & 61.63 & 61.31 & 48.75 & \cellcolor{avgpurple}68.21 \\
    PuRF (Ours)   & \cellcolor{avgblue}\textbf{88.31} & \cellcolor{avgblue}\textbf{86.97}      & \cellcolor{avgblue}\textbf{66.69} & \cellcolor{avgblue}\textbf{66.02} & \cellcolor{avgblue}\textbf{50.84}      &  \cellcolor{avgpurple}\textbf{71.77} \\
    \midrule
    \rowcolor{gray!10}
    \multicolumn{7}{c}{\textbf{\textit{MaPLe}}} \\
    \midrule
    MaPLe~\cite{maple}\textsubscript{\scriptsize CVPR'23} & 85.34 & 84.79 & 62.18 & 62.35 & 48.42 & \cellcolor{avgpurple}68.62 \\
    \midrule
    TPT~\cite{tpt}\textsubscript{\scriptsize NeurIPS'22}   & 85.04 & 83.92 & 63.36 & 63.75 & 48.90 & \cellcolor{avgpurple}69.01 \\
    DiffTPT~\cite{difftpt}\textsubscript{\scriptsize CVPR'23} & 85.15 & 83.78 & 62.93 & 63.14 & 48.81 & \cellcolor{avgpurple}68.76 \\
    RLCF~\cite{rlcf}\textsubscript{\scriptsize ICLR'24}  & 85.35 & 85.28 & 62.84 & 62.90 & 49.37 & \cellcolor{avgpurple}69.15 \\
    ML-TTA~\cite{mltta}\textsubscript{\scriptsize ICLR'25} & 86.40 & 85.69 & 64.75 & 64.86 & 50.21 & \cellcolor{avgpurple}70.38 \\
    \midrule
    TDA~\cite{tda}\textsubscript{\scriptsize CVPR'24}   & 85.76 & 84.15 & 63.25 & 63.64 & 49.55 & \cellcolor{avgpurple}69.27 \\
    DPE~\cite{dpe}\textsubscript{\scriptsize NeurIPS'24}   & 87.55      & 86.73      & 64.80      & 64.38      & 52.39      & \cellcolor{avgpurple}71.17 \\
    BoostAdapter~\cite{boostadapter}\textsubscript{\scriptsize NeurIPS'24} & 85.99      & 84.62      & 63.74      & 63.62      & 50.16      & \cellcolor{avgpurple}69.63 \\
    ReTA~\cite{reta}\textsubscript{\scriptsize ACM MM'25}  & 87.40 & 86.60 & 64.59 & 64.21 & 51.62 & \cellcolor{avgpurple}70.88 \\
    PuRF (Ours)   & \cellcolor{avgblue}\textbf{89.42} & \cellcolor{avgblue}\textbf{88.56} & \cellcolor{avgblue}\textbf{69.31} & \cellcolor{avgblue}\textbf{68.82} & \cellcolor{avgblue}\textbf{52.84} & \cellcolor{avgpurple}\textbf{73.79} \\
    \midrule
    \rowcolor{gray!10}
    \multicolumn{7}{c}{\textbf{\textit{Template}}} \\
    \midrule
    CLIP~\cite{clip}\textsubscript{\scriptsize ICML'21}   & 82.93      & 82.26      & 59.94      & 59.62      & 47.76      & \cellcolor{avgpurple}66.50 \\
    \midrule
    TDA~\cite{tda}\textsubscript{\scriptsize CVPR'24}   & 83.18      & 82.09      & 60.06      &  60.05     &  47.95     & \cellcolor{avgpurple}66.67 \\
    DPE~\cite{dpe}\textsubscript{\scriptsize NeurIPS'24}   & 85.86      & 84.82      & 61.99      & 61.54      & 49.29      & \cellcolor{avgpurple}68.70 \\
    BoostAdapter~\cite{boostadapter}\textsubscript{\scriptsize NeurIPS'24} & 84.85      & 84.06      & 61.82      &  61.79     &  48.64     & \cellcolor{avgpurple}68.23 \\
    ReTA~\cite{reta}\textsubscript{\scriptsize ACM MM'25}  & 87.71      & 86.54      & 63.43      & 62.56      & 48.41      & \cellcolor{avgpurple}69.73 \\
    PuRF (Ours)   & \cellcolor{avgblue}\textbf{88.73} & \cellcolor{avgblue}\textbf{87.75} & \cellcolor{avgblue}\textbf{66.94} & \cellcolor{avgblue}\textbf{66.16} & \cellcolor{avgblue}\textbf{50.47}      & \cellcolor{avgpurple}\textbf{72.01} \\
    \bottomrule
    \end{tabular}%
    }
  \label{tab: 2}%
  \vspace{-0.15cm}
\end{table}%

\begin{table}[!t] 
  \centering
    \caption{Efficiency and performance over five multi-label datasets on a single NVIDIA 3090 GPU (ViT-B/16).}
  \vspace{-0.1cm}
  \resizebox{0.9\columnwidth}{!}{
    \begin{tabular}{lccccc}
    \toprule
    Method 
    & {Inference Speed (FPS)\textbf{ }{\small\textcolor{blue}{$\mathbf{\uparrow}$}}} 
    & {Memory (GB)\textbf{ }{\small\textcolor{blue}{$\mathbf{\downarrow}$}}}
    & {mAP (\%)\textbf{ }{\small\textcolor{blue}{$\mathbf{\uparrow}$}}} 
    & {Gain (\%)\textbf{ }{\small\textcolor{blue}{$\mathbf{\uparrow}$}}} \\
    \midrule
    CLIP~\cite{clip}      & 106.25 & 0.35  & 62.61 &  -    \\
    TPT~\cite{tpt}        &  3.59 & 5.77 & 61.72 & -0.89 \\
    ML-TTA~\cite{mltta}   &  3.05 & 5.77 & 64.80 &  2.19 \\
    DPE~\cite{dpe}        &  4.78 & 7.49  & 67.02 &  4.41 \\
    ReTA~\cite{reta}      &  6.35 & 0.74  & 67.89 &  5.28 \\
    \cellcolor{avgblue}PuRF (Ours) 
                          & \cellcolor{avgblue}5.74   
                          & \cellcolor{avgblue}1.04 
                          & \cellcolor{avgblue}71.48 
                          & \cellcolor{avgblue}\textbf{8.87} \\
    \bottomrule
    \end{tabular}
  }
  % \vspace{-0.28cm}
  \vspace{-0.3cm}
  \label{tab:speed}
\end{table}

\textbf{Implementation Details. }We verify the generalization of our PuRF across diverse CLIP architectures (RN50, RN101, ViT-B/16, ViT-B/32) 
using the default prompt ``\textit{a photo of a}'' and report results in mAP.
% using mAP as the evaluation metric.
We further assess its compatibility with different prompt initializations, including (1) pretrained prompt weights from MaPLe~\cite{maple} and CoOp~\cite{coop}, and (2) dataset-specific templates in CuPL~\cite{cupl} manner.
% We evaluate our method across multiple CLIP architectures with visual encoder backbones including RN50, RN101, ViT-B/16, and ViT-B/32, with mean Average Precision (mAP) as the main evaluation metric.
% For prompt initialization, we consider three settings: (1) the simple prompt template “a photo of a” following ML-TTA, (2) pretrained prompt weights from MaPLe and CoOp, and (3) dataset-customized prompts constructed in the CuPL manner. 
% The number of generated regions is set to 50 for all datasets.
For all datasets, we set the number of generated regions to 50 and fix the temporal decay constant as $\delta = 1000$.
For global filtering, we select the top-$\kappa_g$ labels with $\kappa_g=0.1$.
% while the local region-level mean is computed from the top-5 regions by default. 
% while the local branch applies $K_l{=}5$ for region-level averaging with a cropping scale sampled from [0.5, 0.8] following~\cite{lazsl}.
% We use the AdamW optimizer with a learning rate of 0.001. 
% The batch size is set to 1.
We use 63 augmented visual views, consistent with common practice in prior baselines~\cite{tpt,tda,dpe,boostadapter}.
The loss coefficients are set to $\lambda_1{=}0.2$ and $\lambda_2{=}0.5$.
The cache size $L{=}3$ with three adjacent text embeddings follows~\cite{reta, dpe}. Please refer to Supplementary for more details.
% Unless otherwise specified, all ablation studies are conducted on COCO2014, and more details are in the Supplementary.

% ----------------------------------------------------------------

\subsection{Comparison with State-of-the-Art Methods}
% 可以不写小标题，然后把accuracy和efficiency都放在一块吗？ efficiency对比cache-based的可能会有劣势？ 可以用每张图片的速度来对比？
% \textbf{Results under Various Visual Encoders. }
% \noindent\textbf{Results under Different Prompt Initializations. }
\textbf{Different CLIP Visual Backbones. }As shown in Table~\ref{tab: 1}, PuRF consistently shows superior performance across all datasets and backbones. 
Notably, on ViT-B/32, PuRF improves mAP by 8.56\% over its prompt-based counterpart ML-TTA~\cite{mltta} and 4.05\% over ReTA~\cite{reta} on average across datasets, demonstrating strong generalization under multi-label setting.
% PuRF surpasses the strongest prompt-based counterpart ML-TTA by an average of 7.41\% mAP, and exceeds the best cache-based competitor ReTA by 3.37\% mAP, demonstrating remarkable generalization and fine-grained representation capability under the multi-label setting.
By incorporating purified informative regions with multi-granularity aggregation and alignment, PuRF extracts reliable and complementary cues for more accurate multi-label recognition.
% In addition, TCR further enhances adaptability especially on large-scale datasets,
% and the overall PuRF composition yields a notable 5.53\% gain over ReTA on COCO2014 across backbones, with consistent gains on NUS-WIDE.
In addition, the purified cache enables accurate class-wise calibration and maintains long-term cache adaptability under evolving distributions.
Consequently, PuRF achieves a notable 6.09\% mAP gain over ReTA on the large-scale COCO2014 dataset and consistent improvements on NUS-WIDE across backbones.

% While prompt-based ML-TTA yields limited improvement over CLIP, cache-based methods show stronger adaptation due to efficient feature reuse. 
% Building on this advantage, ReD achieves further gains, reaching 70.34 / 68.88 mAP on ViT-B/16 and ViT-B/32, exceeding ReTA by 2.45 / 3.05 mAP. These results highlight ReD’s robustness across architectures and its superior capability for fine-grained multi-label adaptation.
% ==================================== Table 3 =======================================

\begin{table}[!t]
    \centering
    \caption{Comparison with state-of-the-art cache-based methods across diverse VLM backbones. Bold indicates best results.}
    \vspace{-1mm}
    \setlength{\tabcolsep}{4.2pt}
    \renewcommand{\arraystretch}{0.86}
    \resizebox{0.92\columnwidth}{!}{
    \begin{tabular}{llcccc}
    \toprule
    & Method & VOC2007 & VOC2012 & COCO2017 & Avg. \\
    \midrule
    % \multirow{4}{*}{EVA-02~\cite{eva02}}
    \multirow{4}{*}{\begin{tabular}[c]{@{}l@{}}EVA-02~\cite{eva02} \\[2pt] \footnotesize (ViT-B/16)\end{tabular}}
    & No Adapt. & 84.05 & 83.88 & 57.88 & 75.27 \\
    & DPE~\cite{dpe}  & 87.19 & 86.49 & 61.22 & 78.30 \\
    & ReTA~\cite{reta} & 86.76 & 86.23 & 61.15 & 78.05 \\
    & PuRF (Ours) & \textbf{89.65} & \textbf{88.43} & \textbf{66.59} & \textbf{81.56} \\
    \midrule
    % \multirow{4}{*}{SigLIP2}
    \multirow{4}{*}{\begin{tabular}[c]{@{}l@{}}SigLIP2~\cite{siglip2} \\[2pt] \footnotesize (SO400M/14)\end{tabular}}
    & No Adapt. & 88.35 & 88.09 & 67.32 & 81.25 \\
    & DPE~\cite{dpe}  & 89.40 & 88.98 & 68.65 & 82.34 \\
    & ReTA~\cite{reta} & 88.84 & 88.47 & 67.92 & 81.74 \\
    & PuRF (Ours) & \textbf{91.13} & \textbf{90.34} & \textbf{72.81} & \textbf{84.76} \\
    \midrule
    % \multirow{4}{*}{MetaCLIP2}
    \multirow{4}{*}{\begin{tabular}[c]{@{}l@{}}MetaCLIP2~\cite{metaclip2} \\[2pt] \footnotesize (ViT-H/14)\end{tabular}}
    & No Adapt. & 84.16 & 83.91 & 61.16 & 76.41 \\
    & DPE~\cite{dpe}  & 85.64 & 84.57 & 62.87 & 77.69 \\
    & ReTA~\cite{reta} & 86.30 & 85.22 & 63.41 & 78.31 \\
    & PuRF (Ours) & \textbf{89.59} & \textbf{88.74} & \textbf{70.29} & \textbf{82.87} \\
    \bottomrule
    \end{tabular}}
    % \vspace{-1mm}
    \label{tab:vlms}
\end{table}

\begin{table}[!t]
  \centering
  \caption{Ablation of key components on ViT-B/16.}
  \vspace{-0.02cm}
  \resizebox{0.62\columnwidth}{!}{%
    \begin{tabular}{c|ccc|@{\hskip 1pt}r@{\hskip 1pt}|cc}
    \toprule
          & RP (region)   & EP (cache)  & TR (cache)   &       & VOC2007 & COCO2017 \\
    \midrule
    \multicolumn{7}{@{}c@{}}{}\\[-15pt]
    \midrule
    1     &       &       &       &       & 84.83 & 60.71 \\
    2     & $\checkmark$   &       &       &       & 86.67 & 63.35 \\
    3     &       & $\checkmark$   &       &       & 85.79      & 62.59 \\
    4     & $\checkmark$   & $\checkmark$   &       &       & 87.50 & 64.56 \\
    5     & $\checkmark$   &       & $\checkmark$   &       & 87.36      & 64.12  \\
    6     &       & $\checkmark$   & $\checkmark$   &       & 86.65      &  63.37 \\
    \cellcolor{avgblue}7     & \cellcolor{avgblue}$\checkmark$   & \cellcolor{avgblue}$\checkmark$   & \cellcolor{avgblue}$\checkmark$   &     & \cellcolor{avgblue}\textbf{88.18}\cellcolor{avgblue} & \cellcolor{avgblue}\textbf{65.35}\cellcolor{avgblue} \\
    \bottomrule
    \end{tabular}%
    }
    \vspace{-0.05cm}
  \label{tab:component}%
\end{table}%

\begin{table}[!t]
    % \captionsetup{font=normalsize,labelformat=empty} % 局部设置
    % \captionsetup[subtable]{labelformat=empty}
    \centering
    %======================= 左边 =======================%
    \begin{subtable}{.49\columnwidth}
        \centering
        \captionsetup{font=small, labelformat=empty}
        \vspace{0.1cm}
        {\small
        \caption{\textbf{Table 6:} Ablation on Region Purification (COCO2017, ViT-B/16).}
        \vspace{-0.1cm}
            \resizebox{0.8\columnwidth}{!}{
            \begin{tabular}{lc}
            \toprule
            Region Selection & mAP (\%) \\
              \midrule
             Global-only & 61.67 \\     
             All regions  & 63.46 \\  
            Abs-score selection  & 65.02 \\      
            \cellcolor{avgblue}Rel-score selection (ours)  & \cellcolor{avgblue}\textbf{65.35} \\
            \bottomrule
            \end{tabular}
            }
        }
        \label{tab:logits_comp}
        
    \end{subtable}%
    % \hspace{0.05\columnwidth} % 控制左右间距
    \hspace{0.02\columnwidth} % 可选的微调间距
    %======================= 右边 =======================%
    \begin{subtable}{.45\columnwidth}
        \centering
        \captionsetup{font=small, labelformat=empty}
        \vspace{0.1cm}
        {\small % ← 仅本子表内 tabular 用 9pt
        \caption{\textbf{Table 7:}  Ablation on Cache Purification (COCO2017, ViT-B/16).}
        \vspace{-0.1cm}
        \resizebox{0.96\columnwidth}{!}{
        \begin{tabular}{lc}
              \toprule
              Cache setting & mAP (\%) \\
              \midrule
                Global-only cache & 62.21 \\        
                PuRF (global retrieve) & 64.80 \\      
                PuRF (min-ent) + avg & 64.92 \\     
                \cellcolor{avgblue}PuRF (min-ent) + max (ours) & \cellcolor{avgblue}\textbf{65.35} \\ 
            \bottomrule
            \end{tabular}
            }
        }
        \label{tab:setcomp}
    \end{subtable}%
    \vspace{-0.35cm}
\end{table}

\textbf{Different Prompt Initialization. }In Table~\ref{tab: 2}, PuRF still achieves the best performance under all prompt initialization methods. 
With MaPLe~\cite{maple} initialization, PuRF achieves 73.79\% mAP, outperforming ReTA (70.88\%) and ML-TTA (70.38\%), highlighting its strong adaptability to various prompt initializations.
% Methods with well-trained prompts (\textit{e.g.}, MaPLe~\cite{maple}) perform better, while our PuRF achieves 73.33\% with MaPLe initialization, outperforming ReTA (70.88\%) and ML-TTA (70.38\%), highlighting its strong adaptability to various prompt initializations.
Templates constructed via CuPL~\cite{cupl} offer richer semantics that benefit cache-based methods, from which PuRF further benefits by exploiting purified regional cues for more reliable class-wise predictions, surpassing ReTA by 2.28\% mAP. 
While prompt-based methods are incompatible with such template design and thus fail to effectively exploit its semantic richness.
% It surpasses ReTA and ML-TTA by 3.04 / 4.92 mAP under CoOp, and by 2.47 / 2.97 mAP under MaPLe. Even with the hand-crafted template, ReD maintains strong results (76.62 mAP), demonstrating stable advantages and general applicability across varying prompt designs.

% \begin{table}[!t]
%   \centering
%   \caption{Ablation of key components on ViT-B/16.}
%   \vspace{-0.18cm}
%   \resizebox{0.47\columnwidth}{!}{%
%     \begin{tabular}{c|ccc|@{\hskip 1pt}r@{\hskip 1pt}|cc}
%     \toprule
%           & MCA   & DRPC  & TCR   &       & VOC2007 & COCO2014 \\
%     \midrule
%     \multicolumn{7}{@{}c@{}}{}\\[-15pt]
%     \midrule
%     1     &       &       &       &       & 84.83 & 60.84 \\
%     2     & $\checkmark$   &       &       &       & 86.67 & 63.31 \\
%     3     &       & $\checkmark$   &       &       & 85.79      & 62.47 \\
%     4     & $\checkmark$   & $\checkmark$   &       &       & 87.50 & 65.29 \\
%     5     & $\checkmark$   &       & $\checkmark$   &       & 87.36      & 64.35  \\
%     6     &       & $\checkmark$   & $\checkmark$   &       & 86.65      &  63.96 \\
%     \cellcolor{avgblue}7     & \cellcolor{avgblue}$\checkmark$   & \cellcolor{avgblue}$\checkmark$   & \cellcolor{avgblue}$\checkmark$   &     & \cellcolor{avgblue}\textbf{88.18}\cellcolor{avgblue} & \cellcolor{avgblue}\textbf{66.05}\cellcolor{avgblue} \\
%     \bottomrule
%     \end{tabular}%
%     }
%     \vspace{-0.3cm}
%   \label{tab:component}%
% \end{table}%

\textbf{Efficiency Analysis. }Table~\ref{tab:speed} compares the efficiency and performance of different baselines on a single NVIDIA 3090 GPU.
% We evaluate the efficiency and performance of different baselines, where the speed is measured in frames per second (FPS) on a single NVIDIA 3090 GPU in Table~\ref{tab:speed}. 
Prompt-based methods (\textit{e.g}., \cite{tpt,mltta}) exhibit high latency and memory due to repeated prompt re-initialization and optimization. 
DPE requires expensive visual prototype learning that leads to high memory consumption.
In contrast, our PuRF adopts lightweight text residual learning with tolerable region-level computation, achieving a remarkable 8.87\% mAP gain with low memory cost.
% Our ReD inherits the efficiency of cache-based TTA methods by reusing historical information for lightweight, reset-free optimization, achieving fine-grained multi-label recognition with a remarkable 8.48 mAP gain and a superior balance between efficiency and accuracy.

%%% 新增 %%%
\textbf{Generalization across Diverse VLM Backbones. }To validate the generalization of PuRF beyond the original CLIP, we further evaluate it on three advanced VLM backbones, including EVA-02~\cite{eva02} (ViT-B/16), SigLIP2~\cite{siglip2} (SO400M/14), and MetaCLIP2~\cite{metaclip2} (ViT-H/14).
As shown in Table~\ref{tab:vlms}, PuRF consistently achieves the best performance across all datasets compared with two strong cache-based baselines, yielding average improvements of 3.56\% mAP over DPE~\cite{dpe} and 3.69\% mAP over ReTA~\cite{reta}.
These results demonstrate the effectiveness and generalization of PuRF across multiple VLM architectures.

\subsection{Ablation Studies}
% \subsection{Model Analysis} % 常规的

\textbf{Effectiveness of the Key Components. }In Table~\ref{tab:component}, we analyze the effectiveness of each module in PuRF, including Region Purification (RP), Episodic Purification (EP) and Temporal Refreshing (TR) for cache purification.
% Disentangled Region Prototype Caching (DRPC), and Temporal Cache Refreshing (TCR).
% RP succeeds in exploiting informative regions through region purification, providing complementary fine-grained cues that enable reliable pseudo supervision via global–local consistency for semantic optimization, thereby significantly boosting recognition performance by an average of 2.24\% across 2 datasets.
RP exploits informative regions through purification, providing fine-grained cues that enable reliable pseudo supervision via global-local consistency for semantic optimization, boosting performance by 2.24\% on average across two datasets.
% RP enhances fine-grained alignment between global and local features, enabling more reliable pseudo-supervision and boosting performance by an average of 2.72\% across 2 datasets, which highlights its effectiveness in improving label consistency.
EP improves cache purity by constructing class-specific prototypes from purified regions, enabling more effective cache calibration.
It yields a larger gain on COCO2017 (+1.88\% mAP), where dense label co-occurrence requires finer class-wise discrimination.
% refines the cache by disentangling region-label representations, thereby preserving class-specific semantics more effectively. 
TR offers comparable improvements to EP by enhancing long-term cache adaptability and mitigating cache saturation and early-stage bias.
Together, these components form a complementary design that balances performance and temporal adaptivity in multi-label test-time adaptation.

\textbf{Impact of Region and Cache Purification. }
\tableref{tab:logits_comp} and~\tableref{tab:setcomp} evaluate the impact of purification at region and cache levels. 
% Relative activation-based selection effectively identifies purified informative regions, outperforming both global-only and absolute-score strategies.
% Building cache entries from these purified regions further improves performance, highlighting the importance of clean regional evidence for cache calibration.
% Finally, max aggregation over purified regions achieves the best result, as it emphasizes the most reliable regional cues for multi-label prediction.
In Eq.~\eqref{eq:purified_region_feats}, relative activation-based selection measures the competitiveness of a region \textit{w.r.t.} other class responses, favoring regions with strong class-specific dominance and yielding purer regional evidence.
For cache purification, we compare global-only caching, purified-region caching with global retrieval, and entropy-based filtering with different aggregations.
Caching purified regions improves performance by constructing class-specific prototypes from less ambiguous signals.
Max aggregation further achieves the best results by preserving the most confident class-wise regional activation, preventing dominant responses from being diluted.

% -----------------------------------------------------------------% 

\begin{figure*}[!t]
\hspace*{-0.2cm}  % 调整负值大小，控制左移距离
    % % --- (a) ---
    % \begin{minipage}[t]{0.19\linewidth}
    %     \centering
    %     \includegraphics[width=1.08\linewidth]{figure/Kg_ablation_plot_fix.pdf}
    %     \vspace{1pt}
    %     \centerline{\hspace{18pt}(a)}\medskip
    %     \label{fig:hyperparam_a}
    % \end{minipage}
    % \hfill
    % % --- (b) ---
    % \begin{minipage}[t]{0.19\linewidth}
    %     \centering
    %     \includegraphics[width=1.08\linewidth]{figure/Kl_ablation_plot_fix.pdf}
    %     \vspace{1pt}
    %     \centerline{\hspace{22pt}(b)}\medskip
    %     \label{fig:hyperparam_b}
    % \end{minipage}
    % \hfill
    % % --- (c) ---
    % \begin{minipage}[t]{0.18\linewidth}
    %     \centering
    %     \includegraphics[width=1.1\linewidth]{figure/delta_ablation_plot_2.pdf}
    %     \vspace{1pt}
    %     \centerline{\hspace{10pt}(c)}\medskip
    %     \label{fig:hyperparam_c}
    % \end{minipage}
    % \hfill
    % % --- (d) ---
    % \begin{minipage}[t]{0.18\linewidth}
    %     \centering
    %     \includegraphics[width=1.1\linewidth]{figure/kappa_ablation_plot_2.pdf}
    %     \vspace{1pt}
    %     \centerline{\hspace{6pt}(d)}\medskip
    %     \label{fig:hyperparam_d}
    % \end{minipage}
    % \hfill
    % % --- (e) ---
    % \begin{minipage}[t]{0.21\linewidth}
    %     \centering
    %     \includegraphics[width=1.1\linewidth]{figure/loss_ablation_heatmap.pdf}
    %     \vspace{1pt}
    %     \centerline{\hspace{-5pt}(e)}\medskip
    %     \label{fig:hyperparam_e}
    % \end{minipage}
    % --- (a) ---
    \begin{minipage}[t]{0.3\linewidth}
        \centering
        \includegraphics[width=0.81\linewidth]{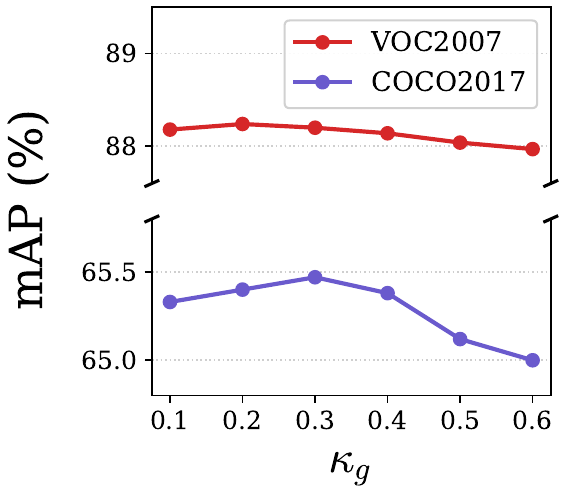}
        \vspace{1pt}
        \centerline{\hspace{18pt}(a)}\medskip
        \label{fig:hyperparam_a}
    \end{minipage}
    \hspace{0.01\linewidth}
    % --- (b) ---
    \begin{minipage}[t]{0.3\linewidth}
        \centering
        % 使用第一张图作为替代
        \includegraphics[width=0.81\linewidth]{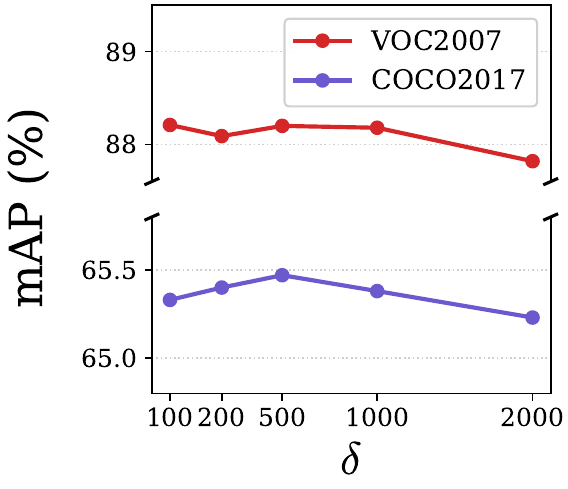} 
        \vspace{1pt}
        \centerline{\hspace{22pt}(b)}\medskip
        \label{fig:hyperparam_b}
    \end{minipage}
    \hspace{0.03\linewidth}
    % --- (c) ---
    \begin{minipage}[t]{0.32\linewidth}
        \centering
        % 保留热力图
        \includegraphics[width=0.92\linewidth]{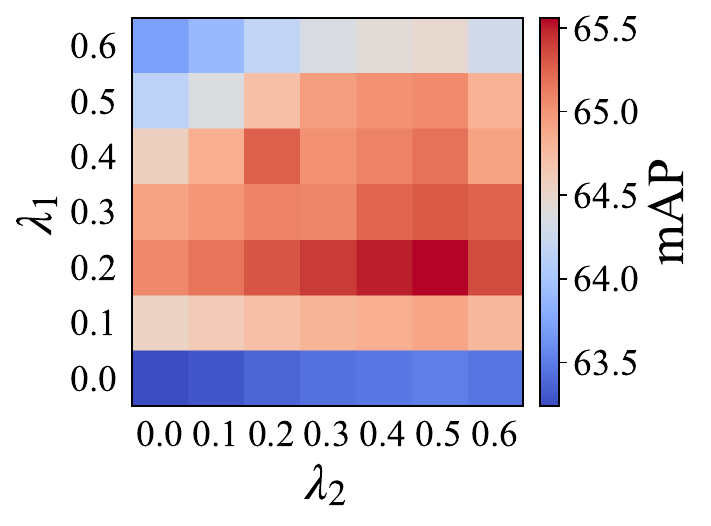} 
        \vspace{1pt}
        \centerline{(c)}\medskip
        \label{fig:hyperparam_c}
    \end{minipage}
    \vspace{-0.35cm}
% \caption{Impact of hyper-parameters on COCO2017. (a) Variation of $K_\mathrm{g}$ in Eq.~\eqref{eq:globalcandidate} . (b) Variation of $K_l$ in Eq.~\eqref{Eq:allscore}. (c) Variation of $\delta$ in Eq.~\eqref{eq:tcr}. (d) Variation of $\kappa$ in Eq.~\eqref{eq:tcr}. (e) Joint impact of $\lambda_1$ and $\lambda_2$ in Eq.~\eqref{all_loss}.}
\caption{Impact of hyper-parameters. (a) Variation of $\kappa_g$ in Eq.~\eqref{eq:globalcandidate}. (b) Variation of $\delta$ in Eq.~\eqref{eq:tcr}. (c) Joint impact of $\lambda_1$ and $\lambda_2$ in Eq.~\eqref{all_loss} on COCO2017.}
    \label{fig: hyperparam}
    \vspace{-0.39cm}
\end{figure*}

% \begin{figure}[!t]
% \centering
%     \vspace{-0.05cm}
% \centerline{\includegraphics[width=6.5cm]{figure/loss_ablation_combine_bar_4.pdf}} 
%     \vspace{-0.4cm}
% \caption{Analysis of different loss designs, including multi-label objectives and ablations by removing each component.}
%     \label{fig: lossdesign}
%     \vspace{-0.3cm}
% \end{figure}

% we find that setting $x$ within 9–15 better captures label co-occurrence while avoiding excessive noise.
% Using only global or local logits yields suboptimal results due to incomplete spatial or contextual representation.
% Combining both improves performance by capturing complementary cues, while introducing the cache term further enhances calibration and robustness, achieving the best 63.61 mAP.
% As shown in Table~\ref{tab:pseudo-consistency}, enforcing consistency between global and local predictions through the intersection-based voting strategy ($\mathcal{G}\cap\mathcal{R}$) produces the most reliable pseudo-labels.
% This confirms that consistency-guided filtering effectively suppresses noisy predictions and strengthens multi-label alignment within MCA.

\begin{figure}[!t]
\centering
    % \vspace{-0.05cm}
\centerline{\includegraphics[width=9.3cm]{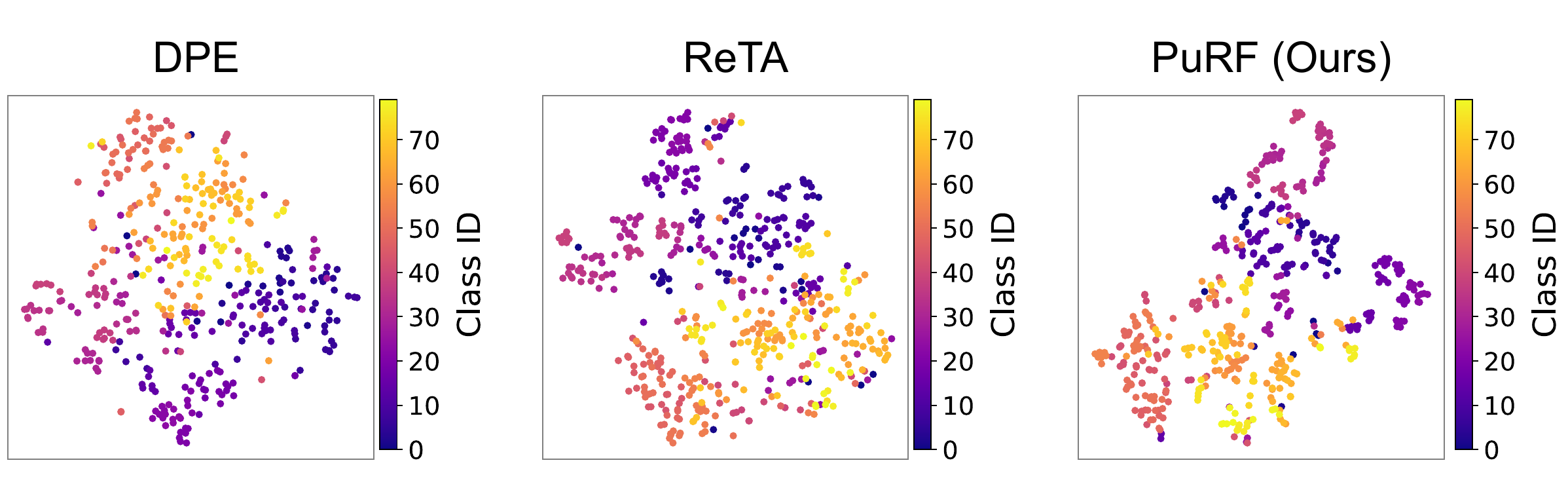}} 
    \vspace{-0.1cm}
\caption{t-SNE visualization of cached visual features on COCO2017. The capacity of all caches is set to 7, and the visualization is taken at the 2000-th adaptation step.}
    \label{fig: tsne}
    \vspace{-0.2cm}
\end{figure}

\textbf{Hyperparameter Sensitivity Analysis.} Here, we analyze four hyperparameters in PuRF:
$\kappa_g$ for selecting the top-$\kappa_g$ classes from global predictions,
$\delta$ for the refreshing onset and entropy penalization scale,
and $\lambda_1,\lambda_2$ for balancing the trade-off between different loss terms.
% As shown in Fig.~\ref{fig: hyperparam}(c),$\delta{=}10^3$-$2{\times}10^3$ yields the best performance, striking a balance between between early and late refreshing. 
As shown in Fig.~\ref{fig: hyperparam}\hyperref[fig: hyperparam]{(a)} and \hyperref[fig: hyperparam]{(b)}, smaller $\kappa_g$ suppresses noisy candidate classes, while $\delta{=}500$-$1000$ strikes a balance between early and delayed refreshing, with stable performance across a wide range.
In Fig.~\ref{fig: hyperparam}\hyperref[fig: hyperparam]{(c)}, $\lambda_1=0.2$ and $\lambda_2=0.5$ achieve the optimal performance.
% achieves the optimal result in regulating the temporal update strength of cached samples.

\begin{figure}[!t]
\centering
    % \vspace{-0.05cm}
\centerline{\includegraphics[width=8.9cm]{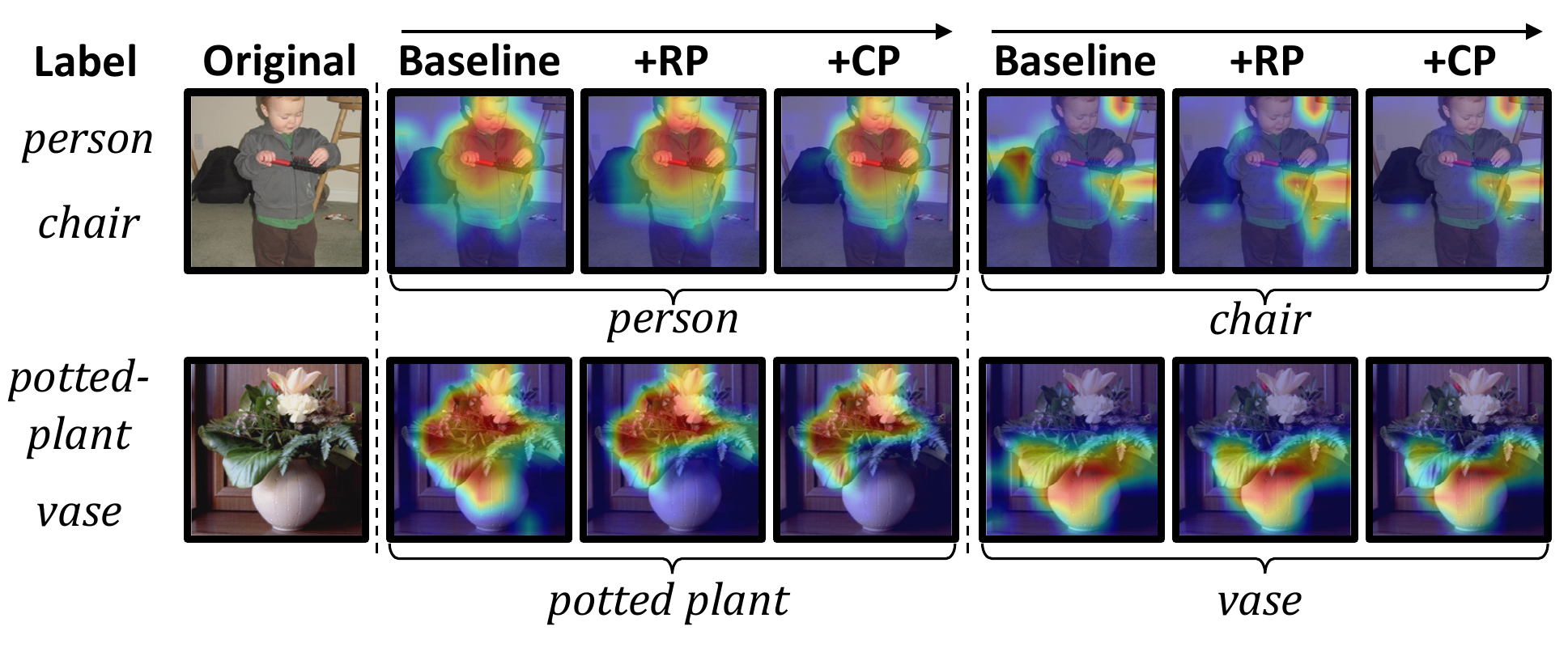}} 
    \vspace{-0.1cm}
\caption{Visualization of class activation maps for region and cache purification.}
    \label{fig: cam}
    \vspace{-0.1cm}
\end{figure}

\subsection{Visualization}

% \textbf{t-SNE Visualization.}
% Figure 4 provides qualitative comparisons of cached representations and instances among DPE, ReTA, and our ReD. In (a), the t-SNE visualization shows that ReD produces more compact and clearly separated clusters, indicating a purer and more discriminative cache distribution than prior methods. In (b), we visualize representative cached images for three semantic concepts—cat, tree, and nighttime. ReD preserves category-consistent and visually reliable samples, while DPE and ReTA tend to include ambiguous or semantically mixed instances. These results demonstrate that ReD effectively maintains a cleaner cache by filtering out noisy samples and capturing fine-grained semantic cues essential for multi-label recognition.
Fig.~\ref{fig: tsne} presents the t-SNE~\cite{tsne} visualization of cached features for DPE~\cite{dpe}, ReTA~\cite{reta}, and our PuRF.
% Compared to the other two methods, ReD forms more compact and clearly separated feature clusters with improved discriminability.
% Although DPE leverages visual-text prototype jointly evolving and ReTA constructs a more reliable vision cache, both methods are originally designed for single-label classification. 
% When applied to multi-label scenarios, the strong label co-occurrence inevitably causes severe feature entanglement, which is visually reflected by heavily overlapping feature clusters across different categories.
% In contrast, the region-disentanglement design in our ReD forms more compact and clearly separated feature clusters with improved discriminability, as it effectively decouples region-specific semantics.
% Although DPE and ReTA improve prototype quality, they are tailored for single-label recognition, where label co-occurrence in multi-label settings causes severe feature entanglement with noisy overlap. 
DPE and ReTA improve cache prototype quality for single-label recognition, but rely on global representations that entangle multiple labels, resulting in noisy and ambiguous cache entries in multi-label settings.
% improve prototype quality, they are tailored for single-label recognition and rely on global representations that entangle co-occurring labels, resulting in noisy and ambiguous cache entries in multi-label settings.
In contrast, PuRF produces class-specific representations with well-separated clusters and strong inter-class discriminability.
Fig.~\ref{fig: cam} illustrates the role of RP and CP. 
RP enhances fine-grained evidence aggregation by exploiting purified regions and enforcing global-local consistency for effective semantic optimization.
EP leverages purified region-level representations to refine class-specific focus with more discriminative features, yielding improved cache calibration.

\section{Conclusion}
% In this paper, we address the challenges of multi-label test-time adaptation and propose ReD, a novel cache-based method tailored for this task.
In this work, we focus on the largely unexplored problem of multi-label test-time adaptation.
% , where feature entanglement among co-occurring labels hinders adaptation performance.
We propose PuRF, a purification-driven cache-based method that addresses the limitations of global representations in multi-label TTA, where label entanglement introduces bias and ambiguity in predictions and cache.
% , where entangled labels introduce bias and ambiguity in predictions and cache.
PuRF performs region and cache purification to effectively exploit informative and reliable regions, improving recognition accuracy and discriminative cache calibration.
% Multi-granularity Consistency Alignment (MCA) for reliable pseudo-labels generation, Disentangled Region Prototype Caching (DRPC) for discriminative class-wise prototypes, and Temporal Cache Refreshing (TCR) to mitigate cache saturation. 
% 这句应该主要说entanglement地问题！ 
% Unlike existing prompt-based or cache-based approaches that suffer from inefficiency or feature entanglement, ReD introduces a region-disentangled mechanism composed of three key modules: Multi-granularity Consistency Alignment (MCA) for reliable pseudo-label generation, Disentangled Region Prototype Caching (DRPC) for purer class representations, and Temporal Cache Refreshing (TCR) to mitigate cache saturation. 
Extensive experiments on five benchmarks verify its consistent superiority over existing methods, achieving state-of-the-art results and establishing it as a strong baseline for future multi-label adaptation research.

%%%%%%%%%%%%%%%%%%%%%%%%%%%%%%%%%%%%%%%%%%%%%%%%%%

% \clearpage\mbox{}Page \thepage\ of the manuscript.
% \clearpage\mbox{}Page \thepage\ of the manuscript.
% \clearpage\mbox{}Page \thepage\ of the manuscript.
% \clearpage\mbox{}Page \thepage\ of the manuscript.
% \clearpage\mbox{}Page \thepage\ of the manuscript. This is the last page.
% Now we have reached the maximum length of an ECCV \ECCVyear{} submission (excluding references and acknowledgements).
% References should start immediately after the main text, but can continue past p.\ 14 if needed. 
% \clearpage  % TODO FINAL: This \clearpage needs to be removed from both review and camera-ready versions.

\section*{Acknowledgements}
This work was supported by National Natural Science Foundation of China (Nos. 62271281, 62525103, 62441235) and Beijing Natural Science Foundation (L247026).

\bibliographystyle{splncs04}
\bibliography{main}
\end{document}